\documentclass[runningheads]{llncs}
\usepackage[T1]{fontenc}
\usepackage{graphicx}
\usepackage{amsmath,amssymb} 

\usepackage[numbers,sort&compress]{natbib}
\usepackage{booktabs}

\usepackage{rotating}

\usepackage{enumitem}

\usepackage{pifont}
\usepackage{xspace}
\usepackage{multirow}
\usepackage{color}
\usepackage[dvipsnames]{xcolor}
\usepackage{tabularx, colortbl}
\usepackage{hyperref}
\hypersetup{
	colorlinks=true,
	linkcolor=black,
	filecolor=black,      
	urlcolor=black,
	citecolor=black
}
\renewcommand{\b}{\boldsymbol}
\newcommand{\m}{\mathcal}
\DeclareMathOperator*{\argmax}{arg\,max}

\newcommand{\cmark}{\ding{51}\xspace}%
\newcommand{\xmark}{\ding{55}\xspace}%
\definecolor{ood}{rgb}{0.96,1,0.94}
\definecolor{idmiscls}{rgb}{1,0.97,0.96}

\newcolumntype{a}{>{\columncolor{ood}}l}
\newcolumntype{b}{>{\columncolor{idmiscls}}l}
\begin{document}
\pagestyle{headings}
\mainmatter

\title{Augmenting Softmax Information for Selective Classification with Out-of-Distribution Data}
\titlerunning{Augmenting the Softmax for Selective Classification with OOD Data}
\authorrunning{Guoxuan Xia and Christos-Savvas Bouganis}

\author{Guoxuan Xia 
\and 
Christos-Savvas Bouganis}
 \institute{Imperial College London \\ \texttt{\{g.xia21,christos-savvas.bouganis\}@imperial.ac.uk}}

\maketitle

\begin{abstract}
Detecting out-of-distribution (OOD) data is a task that is receiving an increasing amount of research attention in the domain of deep learning for computer vision. However, the performance of detection methods is generally evaluated on the task in isolation, rather than also considering potential downstream tasks in tandem. In this work, we examine selective classification in the presence of OOD data (SCOD). That is to say, the motivation for detecting OOD samples is to reject them so their impact on the quality of predictions is reduced. We show under this task specification, that existing post-hoc methods perform quite differently compared to when evaluated only on OOD detection. This is because it is no longer an issue to conflate in-distribution (ID) data with OOD data \textit{if the ID data is going to be misclassified}. However, the conflation within ID data of correct and incorrect predictions becomes undesirable.  
We also propose a novel method for SCOD, Softmax Information Retaining Combination (SIRC), that augments softmax-based confidence scores with feature-agnostic information
such that their ability to identify OOD samples is improved without sacrificing separation between correct and incorrect ID predictions. Experiments on a wide variety of ImageNet-scale datasets and convolutional neural network architectures show that SIRC is able to consistently match or outperform the baseline for SCOD, whilst existing OOD detection methods fail to do so.
Code is available at \url{https://github.com/Guoxoug/SIRC}.
\end{abstract}
\section{Introduction}
Out-of-distribution (OOD) detection \cite{Yang2021GeneralizedOD}, i.e. identifying data samples that do not belong to the training distribution, is a task that is receiving an increasing amount of attention in the domain of deep learning \citep{Liang2018EnhancingTR,Liu2020EnergybasedOD,Du2022VOSLW,Hendrycks2017ABF,Hendrycks2019BenchmarkingNN,Fort2021ExploringTL,Hsu2020GeneralizedOD, Techapanurak_2020_ACCV,Sun2021ReActOD,Wang2022ViMOW,Huang2021MOSTS,Lee2018ASU,Pearce2021UnderstandingSC, Yang2021GeneralizedOD,Zhang2021OnTO, Nalisnick2019DoDG}. The task is often motivated by safety-critical applications, such as healthcare and autonomous driving, where there may be a large cost associated with sending a prediction on OOD data downstream.

However, in spite of a plethora of existing research, there is generally a lack of focus with regards to the specific motivation behind OOD detection in the literature, other than it is often done as part of the pipeline of another primary task, e.g. image classification. As such the task is evaluated in isolation and formulated as binary classification between in-distribution (ID) and OOD data. In this work we consider the question \textit{why exactly do we want to do OOD detection during deployment?} We focus on the problem setting where the primary objective is classification, and we are motivated to detect and then reject OOD data, as predictions on those samples will incur a cost. That is to say the task is selective classification \cite{ElYaniv2010OnTF,Geifman2017SelectiveCF} where OOD data has polluted the input samples. \citet{Kim2021AUB} term this problem setting \textit{unknown detection}. However, we prefer to use Selective Classification in the presence of Out-of-Distribution data (SCOD) as we would like to emphasise the downstream classifier as the objective, and will refer to the task as such in the remainder of the paper.

The \textit{key difference} between this problem setting and OOD detection is that \textit{both} OOD data \textit{and} incorrect predictions on ID data will incur a cost \cite{Kim2021AUB}. It does not matter if we reject an ID sample if it would be incorrectly classified anyway. As such we can view the task as separating correctly predicted ID samples (ID\cmark) from misclassified ID samples (ID\xmark) and OOD samples. This reveals a potential blind spot in designing approaches solely for OOD detection, as the cost of ID misclassifications is ignored. The \textit{key contributions} of this work are:

\begin{enumerate}
    \item Building on initial results from \citep{Kim2021AUB} that show poor SCOD performance for existing methods designed for OOD detection, we show novel insight into the behaviour of different post-hoc (after-training) detection methods for the task of SCOD. Improved OOD detection often comes directly at the expense of SCOD performance. Moreover, the relative SCOD performance of different methods varies with the proportion of OOD data found in the test distribution, the relative cost of accepting ID\xmark vs OOD, as well as the distribution from which the OOD data samples are drawn.
    \item We propose a novel method, targeting SCOD, Softmax Information Retaining Combination (SIRC), that aims to improve the OOD|ID\cmark separation of softmax-based methods, whilst retaining their ability to identify ID\xmark. It consistently outperforms or matches the baseline maximum softmax probability (MSP) approach over a wide variety of OOD datasets and convolutional neural network (CNN) architectures, unlike existing OOD detection methods.  
\end{enumerate}
\section{Preliminaries}\label{prelim}

\subsubsection{Neural Network Classifier}
For a $K$-class classification problem we learn the parameters $\b \theta$ of a discriminative model $P(y|\b x;\b \theta)$ over labels $y \in \m Y = \{\omega_k\}_{k=1}^K$ given inputs $\b x \in \m X = \mathbb R^D$, using finite training dataset $\m D_\text{tr} = \{y^{(n)},\b x^{(n)}\}_{n=1}^{N}$ sampled independently from true joint data distribution $p_\text{tr}(y,\b x)$. 
This is done in order to make predictions $\hat y$ given new inputs $\b x^* \sim p_\text{tr}(\b x)$ with unknown labels,
\begin{equation}\label{classifier}
    \hat y = f(\b x^*) = \argmax_\omega P(\omega|\b x^*;\b \theta)~,
\end{equation}
where $f$ refers to the classifier function. In our case, the parameters $\b \theta$ belong to a deep neural network with categorical softmax output $\b \pi \in [0,1]^K$,
\begin{equation}\label{soft}
     P(\omega_i|\b x;\b \theta) = \pi_i(\b x;\b \theta) = \exp v_i(\b x)/\sum_{k=1}^K \exp v_k(\b x)~,
\end{equation}
where the logits $\b v  = \b W \b z + \b b \quad(\in \mathbb R^K)$ are the output of the final fully-connected layer with weights $\b W \in \mathbb R^{K\times L}$, bias $\b b \in  \mathbb R^K$, and final hidden layer features $\b z \in \mathbb R^L$ as inputs. Typically $\b \theta$ are learnt by minimising the cross entropy loss, such that the model approximates the true conditional distribution $P_\text{tr}(y|\b x)$,
\begin{align}\label{ce}
    \m L_\text{CE}(\b \theta) &= -\frac{1}{N}\sum_{n=1}^{N}\sum_{k=1}^K \delta(y^{(n)}, \omega_k)\log P(\omega_k|\b x^{(n)};\b \theta) \\
    &\approx -\mathbb E_{p_\text{tr}(\b x)}\left[\sum_{k=1}^K P_\text{tr}(\omega_k|\b x)\log P(\omega_k|\b x;\b \theta)\right] = \mathbb{E}_{p_\text{tr}}\left[\text{KL}\left[P_\text{tr}||P_{\b\theta}\right]\right] + A~, \nonumber 
\end{align}
where $\delta(\cdot,\cdot)$ is the Kronecker delta, $A$ is a constant with respect to $\b \theta$ and KL$[\cdot||\cdot]$ is the Kullback–Leibler divergence.
\subsubsection{Selective Classification} A selective classifier \cite{ElYaniv2010OnTF} can be formulated as a pair of functions, the aforementioned classifier $f(\b x)$ (in our case given by Eq. \ref{classifier}) that produces a prediction $\hat y$, and a binary rejection function 
\begin{equation}
    g(\b x;t) = \begin{cases}
    0\text{ (reject prediction)}, &\text{if }S(\b x) < t\\
    1\text{ (accept prediction)}, &\text{if }S(\b x) \geq t~,
    \end{cases}
\end{equation}
where $t$ is an operating threshold and $S$ is a scoring function which is typically a measure of predictive confidence (or $-S$ measures uncertainty). Intuitively, a selective classifier chooses to reject if it is uncertain about a prediction.
\subsubsection{Problem Setting} We consider a scenario where, during deployment, classifier inputs $\b x^*$ may be drawn from either the training distribution $p_\text{tr}(\b x)$ (ID) or another distribution $p_\text{OOD}(\b x)$ (OOD). That is to say,
\begin{equation}\label{mix}
    \b x^* \sim p_\text{mix}(\b x), \quad p_\text{mix}(\b x) = \alpha p_\text{tr}(\b x) + (1-\alpha)p_\text{OOD}(\b x)~,
\end{equation}
where $\alpha \in [0,1]$ reflects the proportion of ID to OOD data found in the wild. Here ``Out-of-Distribution'' inputs are defined as those drawn from a distribution with label space that does not intersect with the training label space $\m Y$ \cite{Yang2021GeneralizedOD}. For example, an image of a car is considered OOD for a CNN classifier trained to discriminate between different types of pets.

We now define the predictive loss on an accepted sample as
\begin{equation}\label{loss}
    \m L_\text{pred}(f(\b x^*)) = \begin{cases}
    0, &\text{if } f(\b x^*) = y^*, \quad y^*, \b x^* \sim p_\text{tr}(y,\b x)  \quad (\text{ID\cmark})\\
    \beta, &\text{if } f(\b x^*) \neq y^*, \quad y^*, \b x^* \sim p_\text{tr}(y,\b x)  \quad (\text{ID\xmark})\\
    1-\beta, &\text{if } \b x^* \sim p_\text{OOD}(\b x)  \quad (\text{OOD})~,
    \end{cases}
\end{equation}
where $\beta \in [0,1]$, and define the selective risk as in \cite{Geifman2017SelectiveCF},
\begin{equation}\label{risk}
    R(f,g;t) = \frac{\mathbb E_{p_\text{mix}(\b x)}[g(\b x;t)\m L_\text{pred}(f(\b x))]}{\mathbb E_{p_\text{mix}(\b x)}[g(\b x;t)]}~,
\end{equation}
which is the average loss of the accepted samples. 
We are only concerned with the relative cost of ID\xmark and OOD samples, so we use a single parameter $\beta$.

The objective is to find a classifier and rejection function $(f,g)$ that minimise $R(f, g;t)$ for some given setting of $t$. We focus on comparing post-hoc (after-training) methods in this work, where $g$ or equivalently $S$ is varied with $f$ fixed. This removes confounding factors that may arise from the interactions of different training-based and post-hoc methods, as they can often be freely combined. 
In practice, both $\alpha$ and $\beta$ will depend on the deployment scenario. However, whilst $\beta$ can be set freely by the practitioner, $\alpha$ is outside of the practitioner's control and their knowledge of it is likely to be very limited.
\begin{figure}[t]
    \centering
    \includegraphics[width=.6\linewidth]{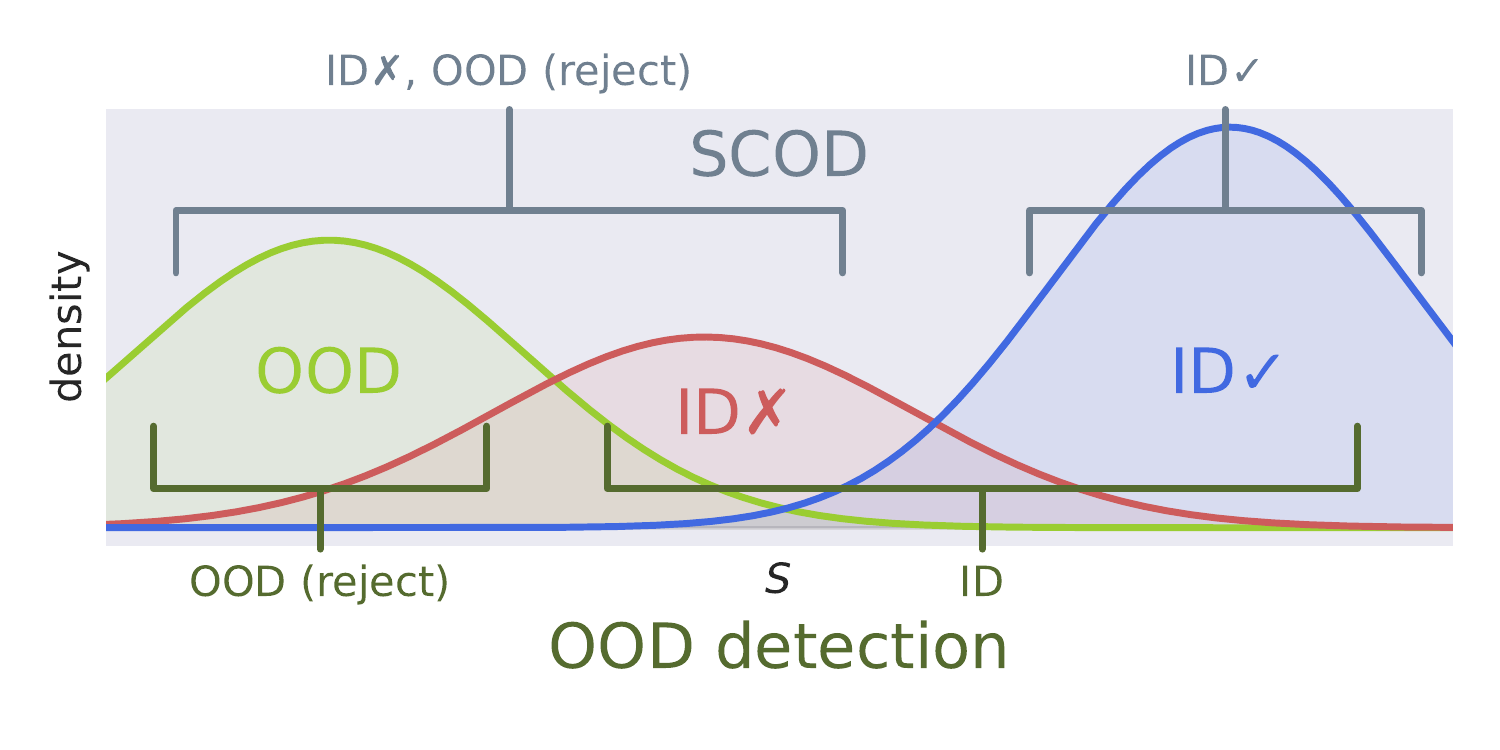}
    \caption{Illustrative sketch showing how SCOD differs to OOD detection. Densities of \textcolor{LimeGreen}{OOD} samples, misclassifications (\textcolor{BrickRed}{ID\xmark}) and correct predictions (\textcolor{NavyBlue}{ID\cmark}) are shown with respect to confidence score $S$. For OOD detection the aim is to separate \textcolor{LimeGreen}{OOD}|\textcolor{BrickRed}{ID\xmark}\textcolor{NavyBlue}{ID\cmark}, whilst for SCOD the data is grouped as \textcolor{LimeGreen}{OOD}\textcolor{BrickRed}{ID\xmark}|\textcolor{NavyBlue}{ID\cmark}.}
    \label{fig:illust}
\end{figure}

It is worth contrasting the SCOD problem setting with OOD detection. SCOD aims to separate OOD, ID\xmark|ID\cmark, whilst for OOD detection the data is grouped as OOD|ID\xmark, ID\cmark (see Fig. \ref{fig:illust}). We note that previous work \cite{kendalunc,Malinin2018PredictiveUE, Malinin2020EnsembleDD, Mukhoti2021DeterministicNN, Pearce2021UnderstandingSC} refer to different types of predictive uncertainty, namely aleatoric and epistemic. The former arises from uncertainty inherent in the data (i.e. the true conditional distribution $P_\text{tr}(y|\b x)$) and as such is irreducible, whilst the latter can be reduced by having the model learn from additional data. Typically, it is argued that it is useful to distinguish these types of uncertainty at prediction time. For example, epistemic uncertainty should be an indicator of whether a test input $\b x^*$ is OOD, whilst aleatoric uncertainty should reflect the level of class ambiguity of an ID input. An interesting result within our problem setting is that the conflation of these different types of uncertainties may not be an issue, as there is no need to separate ID\xmark from OOD, as both should be rejected.

\section{OOD Detectors Applied to SCOD}\label{OODbad}
As the explicit objective of OOD detection is different to SCOD, it is of interest to understand how existing detection methods behave for SCOD. Previous work \cite{Kim2021AUB} has empirically shown that some existing OOD detection approaches perform worse, and in this section we shed additional light as to why this is the case.
\subsubsection{Improving Performance: OOD Detection vs SCOD}
\begin{figure}[t]
    \centering
    \includegraphics[width=\linewidth]{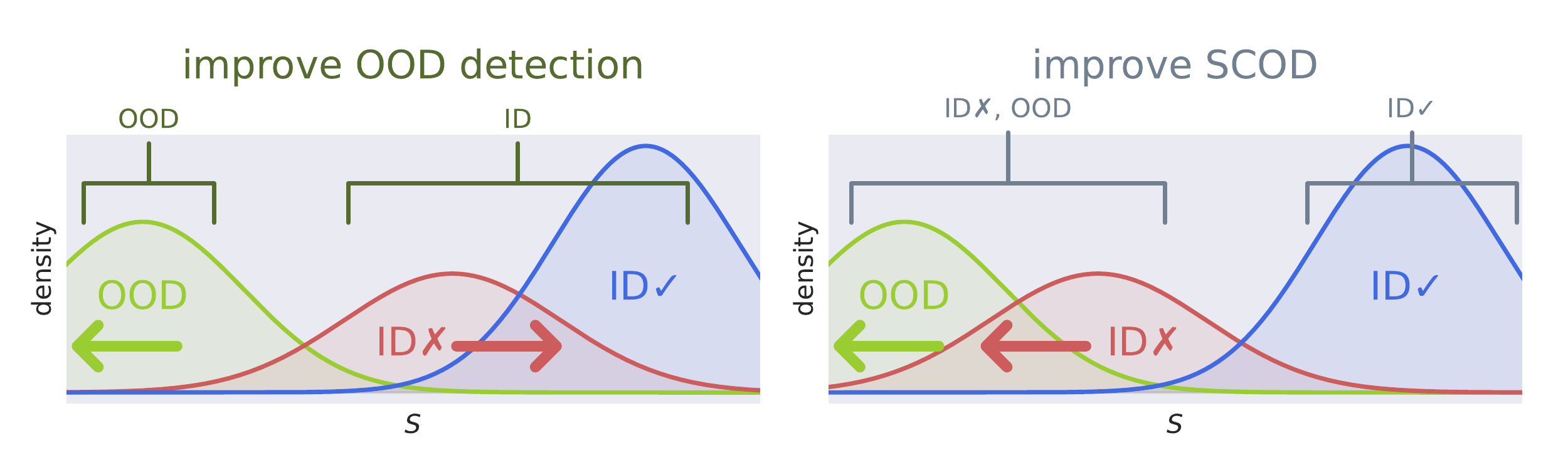}
    \caption{Illustrations of how a detection method can improve over a baseline. \textcolor{OliveGreen}{Left}: For OOD detection we can either have \textcolor{LimeGreen}{OOD} further away from \textcolor{NavyBlue}{ID\cmark} or \textcolor{BrickRed}{ID\xmark} closer to \textcolor{NavyBlue}{ID\cmark}. \textcolor{CadetBlue}{Right}: For SCOD we want both \textcolor{LimeGreen}{OOD} and \textcolor{BrickRed}{ID\xmark} to be further away from \textcolor{NavyBlue}{ID\cmark}. Thus, we can see how improving OOD detection may in fact be at odds with SCOD.}
    \label{fig:improve}
\end{figure}
In order to build an intuition, we can consider, qualitatively, how detection methods can improve performance over a baseline, with respect to the distributions of OOD and ID\xmark relative to ID\cmark. This is illustrated in Fig. \ref{fig:improve}. For OOD detection the objective is to better separate the distributions of ID and OOD data. Thus, we can either find a confidence score $S$ that, compared to the baseline, has OOD distributed further away from ID\cmark, and/or has ID\xmark distributed closer to ID\cmark. In comparison, for SCOD, we want both OOD and ID\xmark to be distributed further away from ID\cmark than the baseline. Thus there is a conflict between the two tasks as, for ID\xmark, the desired behaviour of confidence score $S$ will be different.

\subsubsection{Existing Approaches Sacrifice SCOD by Conflating ID\cmark and ID\xmark}
Considering post-hoc methods, the baseline confidence score $S$ used is Maximum Softmax Probability (MSP) \cite{Hendrycks2017ABF}. Improvements in OOD detection are often achieved by moving away from the softmax $\b \pi$ in order to better capture the differences between ID and OOD data. Energy \cite{Liu2020EnergybasedOD} and Max Logit \cite{Hendrycks2020ScalingOD} consider the logits $\b v$ directly, whereas the Mahalanobis detector \cite{Lee2018ASU} and DDU \cite{Mukhoti2021DeterministicNN} build generative models using Gaussians over the features $\b z$. ViM \cite{Wang2022ViMOW} and Gradnorm \cite{Huang2021OnTI} incorporate class-agnostic, feature-based information into their scores. 

\begin{figure}[t]
    \centering
    \includegraphics[width=\linewidth]{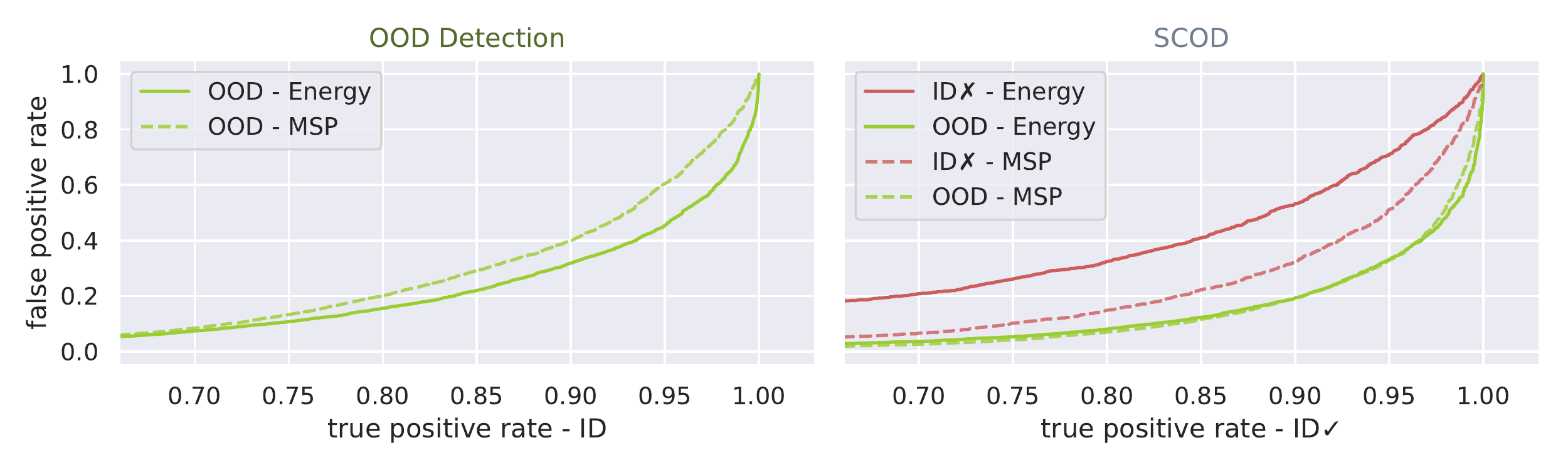}
    \caption{Left: False positive rate (FPR) of \textcolor{LimeGreen}{OOD} samples plotted against true positive rate (TPR) of ID samples. Energy performs better (lower) for OOD detection relative to the MSP baseline. Right: FPR of \textcolor{BrickRed}{ID\xmark} and \textcolor{LimeGreen}{OOD} samples against TPR of \textcolor{NavyBlue}{ID\cmark}. Energy is worse than the baseline at separating \textcolor{BrickRed}{ID\xmark}|\textcolor{NavyBlue}{ID\cmark} and no better for \textcolor{LimeGreen}{OOD}|\textcolor{NavyBlue}{ID\cmark}, meaning it is worse for SCOD. Energy's improved OOD detection performance arises from pushing \textcolor{BrickRed}{ID\xmark} closer to \textcolor{NavyBlue}{ID\cmark}. The ID dataset is ImageNet-200, OOD dataset is iNaturalist and the model is ResNet-50.}
    \label{fig:sep}
\end{figure}

Recall that typically a neural network classifier learns a model $P(y|\b x;\b \theta)$ to approximate the true conditional distribution $P_\text{tr}(y|\b x)$ of the training data (Eqs. \ref{soft},\ref{ce}). As such, scores $S$ extracted from the softmax outputs $\b \pi$ should best reflect how likely a prediction on ID data is going to be correct or not (and this is indeed the case in our experiments in Section \ref{exp}). As the above (post-hoc) OOD detection approaches all involve moving away from the modelled $P(y|\b x;\b \theta)$, we would expect worse separation between ID\xmark and ID\cmark even if overall OOD is better distinguished from ID. Fig. \ref{fig:sep} shows empirically how well different types of data are separated using MSP ($\pi_\text{max}$) and Energy ($\log\sum_k\exp v_k$), by plotting false positive rate (FPR) against true positive rate (TPR). Lower FPR indicates better separation of the negative class away from the positive class. Although Energy has better OOD detection performance compared to MSP, this is actually because the separation between ID\xmark and ID\cmark is much less for Energy, whilst the behaviour of OOD relative to ID\cmark is not meaningfully different to the MSP baseline. Therefore, SCOD performance for Energy is worse in this case. Another way of looking at it would be that for OOD detection, MSP does worse as it conflates ID with OOD, however, this doesn't harm SCOD performance as much, as those ID samples are mostly incorrect anyway. The ID dataset is ImageNet-200 \cite{Kim2021AUB}, OOD dataset is iNaturalist \cite{Huang2021MOSTS} and the model is ResNet-50 \cite{He2016DeepRL}.
\section{Targeting SCOD -- Retaining Softmax Information}\label{sirc}
We would now like to develop an approach that is tailored to the task of SCOD. We have discussed how we expect softmax-based methods, such as MSP, to perform best for distinguishing ID\xmark from ID\cmark, and how existing approaches for OOD detection improve over the baseline, in part, by sacrificing this. As such, to improve over the baseline for SCOD, we will aim to \textit{retain} the ability to separate ID\xmark from ID\cmark whilst \textit{increasing} the separation between OOD and ID\cmark.
\subsubsection{Combining Confidence Scores}
Inspired by Gradnorm \cite{Huang2021OnTI} and ViM \cite{Wang2022ViMOW} we consider the combination of two different confidence scores $S_1, S_2$. We shall consider $S_1$ our primary score, which we wish to augment by incorporating $S_2$. For $S_1$ we investigate scores that are strong for selective classification on ID data, but are also capable of detecting OOD data -- MSP and (the negative of) softmax entropy, $(-)\m H[\b \pi]$. For $S_2$, the score should be useful \textit{in addition} to $S_1$ in determining whether data is OOD or not. We should consider scores that capture different information about OOD data to the post-softmax $S_1$ if we want to improve OOD|ID\cmark.  We choose to examine the $l_1$-norm of the feature vector $||\b z||_1$ from \cite{Huang2021OnTI} and the negative of the Residual\footnote{
$\b z^{P^\bot}$ is the component of the feature vector that lies outside of a principle subspace calculated using ID data. For more details see \citet{Wang2022ViMOW}'s paper.
} score $-||\b z^{P^\bot}||_2$ from \cite{Wang2022ViMOW} as these scores capture class-agnostic information at the feature level. Note that although $||\b z||_1$ and Residual have previously been shown to be useful for OOD detection in \cite{Huang2021OnTI, Wang2022ViMOW}, we do not expect them  to be useful for identifying misclassifications. They are separate from the classification layer defined by $(\b W,\b b)$, so they are far removed from the categorical $P(y|\b x;\b \theta)$ modelled by the softmax.
\subsubsection{Softmax Information Retaining Combination (SIRC)}
We want to create a combined confidence score $C(S_1,S_2)$ that retains $S_1$'s ability to distinguish ID\xmark|ID\cmark but is also able to incorporate $S_2$ in order to augment OOD|ID\cmark. We develop our approach based on the following set of \textit{assumptions}:
\begin{itemize}
    \item $S_1$ will be higher for ID\cmark and lower for ID\xmark and OOD.
    \item $S_1$ is bounded by maximum value $S_1^\text{max}$. \footnote{This holds for our chosen $S_1$ of $\pi_\text{max}$ and $-\m H$.}
    \item $S_2$ is unable to distinguish ID\xmark|ID\cmark, but is lower for OOD compared to ID.
    \item $S_2$ is useful in addition to $S_1$ for separating OOD|ID.
\end{itemize}

\begin{figure}[t]
    \centering
    \includegraphics[width=\linewidth]{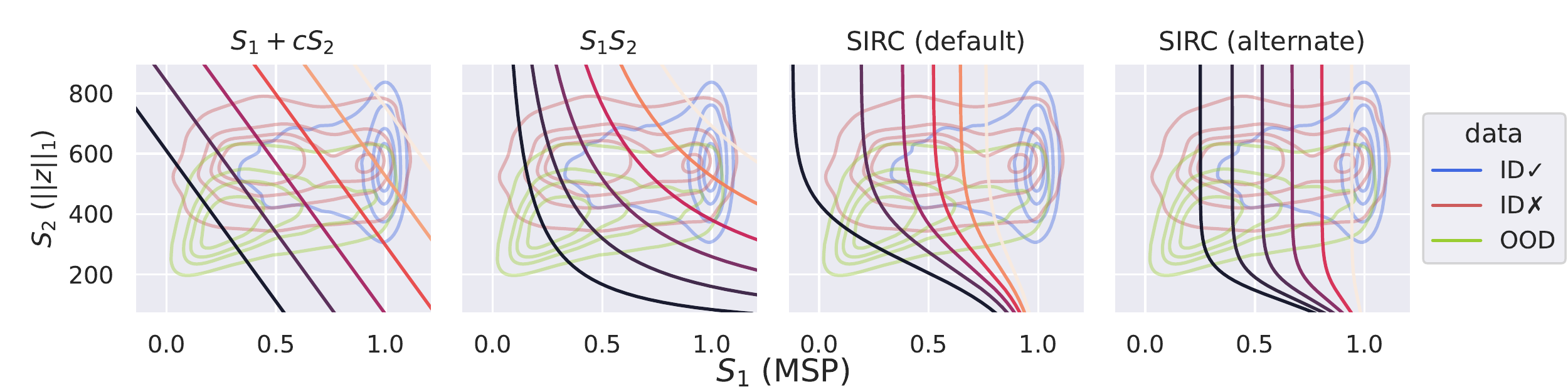}
    \caption{Comparison of different methods of combining confidence scores $S_1,S_2$ for SCOD. \textcolor{LimeGreen}{OOD}, \textcolor{BrickRed}{ID\xmark} and \textcolor{NavyBlue}{ID\cmark} distributions are displayed using kernel density estimate contours. Graded contours for the different combination methods are then overlayed (lighter means higher combined score). We see that our method, SIRC (centre right) is able to better retain \textcolor{BrickRed}{ID\xmark}|\textcolor{NavyBlue}{ID\cmark} whilst improving \textcolor{LimeGreen}{OOD}|\textcolor{NavyBlue}{ID\cmark}. An alternate parameter setting for SIRC, with a stricter adherence to $S_1$, is also shown (far right). The ID dataset is ImageNet-200, the OOD dataset iNaturalist and the model ResNet-50. SIRC parameters are found using ID training data; the plotted distributions are test data.}
    \label{fig:combs}
\end{figure}

We propose to combine $S_1$ and $S_2$ using
\begin{equation}\label{comb}
    C(S_1,S_2) = -(S^{\max}_1-S_1)\left(1+\exp(-b[S_2-a])\right)~,\footnote{To avoid overflow this is implemented using the \texttt{logaddexp} function in PyTorch \cite{torch}.}
\end{equation}
where $a,b$ are parameters chosen by the practitioner. The idea is for the accept/reject decision boundary of $C$ to be in the shape of a sigmoid on the $(S_1,S_2)$-plane (See Fig. \ref{fig:combs}). As such the behaviour of only using the softmax-based $S_1$ is recovered for ID\xmark|ID\cmark as $S_2$ is increased, as the decision boundary tends to a vertical line. However, $S_2$ is considered increasingly important as it is decreased, allowing for improved OOD|ID\cmark. We term this approach Softmax Information Retaining Combination (SIRC).

The parameters $a,b$ allow the method to be adjusted to different distributional properties of $S_2$. Rearranging Eq. \ref{comb},
\begin{equation}
    S_1 = S_1^\text{max} + C/[1+\exp(-b[S_2-a])]~,
\end{equation}
we see that $a$ controls the vertical placement of the sigmoid, and $b$ the sensitivity of the sigmoid to $S_2$. We use the empirical mean and standard deviation of $S_2$, $\mu_{S_2}, \sigma_{S_2}$ on ID data (training or validation) to set the parameters. 
We choose $a = \mu_{S_2}-3\sigma_{S_2}$ so the centre of the sigmoid is below the ID distribution of $S_2$, and we set $b=1/\sigma_{S_2}$, to match the ID variations of $S_2$.  Note that other parameter settings are possible, and practitioners are free to tune $a,b$ however they see fit (on ID data), but we find the above approach to be empirically effective.

Fig. \ref{fig:combs} compares different methods of combination by plotting ID\cmark, ID\xmark and OOD data densities on the $(S_1,S_2)$-plane. Other than SIRC we consider the combination methods used in ViM, $C=S_1 + cS_2$, where $c$ is a user set parameter, and in Gradnorm, $C=S_1 S_2$. The overlayed contours of $C$ represent decision boundaries for values of $t$. We see that the linear decision boundary of $C=S_1 + cS_2$ must trade-off significant performance in ID\xmark|ID\cmark in order to gain OOD|ID\cmark (through varying $c$), whilst $C=S_1 S_2$ sacrifices the ability to separate ID\xmark|ID\cmark well for higher values of $S_1$. We also note that $C=S_1S_2$ is not robust to different ID means of $S_2$. For example, arbitrarily adding a constant $D$ to $S_2$ will completely change the behaviour of the combined score. On the other hand, SIRC is designed to be robust to this sort of variation between different $S_2$. Fig. \ref{fig:combs} also shows an alternative parameter setting for SIRC, where $a$ is lower and $b$ is higher. Here more of the behaviour of only using $S_1$ is preserved, but $S_2$ contributes less. It is also empirically observable that the assumption that $S_2$ (in this case $||\b z||_1$) is not useful for distinguishing ID\cmark from ID\xmark holds, and in practice this can be verified on ID validation data when selecting $S_2$.

We also note that although we have chosen specific $S_1,S_2$ in this work, SIRC can be applied to any $S$ that satisfy the above assumptions. As such it has the potential to improve beyond the results we present, given better individual $S$.
\section{Experimental Results}\label{exp}
We present experiments across a range of CNN architectures and ImageNet-scale OOD datasets. Extended results can be found in Appendix \ref{results}.
\subsubsection{Data, Models and Training}
For our ID dataset we use ImageNet-200 \cite{Kim2021AUB}, which contains a subset of 200 ImageNet-1k \cite{Russakovsky2015ImageNetLS} classes. It has separate training, validation and test sets.
We use a variety of OOD datasets for our evaluation that display a wide range of semantics and difficulty in being identified. Near-ImageNet-200 (Near-IN-200) \cite{Kim2021AUB} is constructed from remaining ImageNet-1k classes semantically similar to ImageNet-200, so it is especially challenging to detect. Caltech-45 \cite{Kim2021AUB} is a subset of the Caltech-256 \cite{Griffin2007Caltech256OC} dataset with non-overlapping classes to ImageNet-200. Openimage-O \cite{Wang2022ViMOW} is a subset of the Open Images V3 \cite{openimages} dataset selected to be OOD with respect to ImageNet-1k. 
iNaturalist \cite{Huang2021MOSTS} and Textures \cite{Wang2022ViMOW} are the same for their respective datasets \cite{inat, cimpoi14describing}. Colorectal \cite{Kather2016MulticlassTA} is a collection of histological images of human colorectal cancer, whilst Colonoscopy is a dataset of frames taken from colonoscopic video of gastrointestinal lesions \cite{Mesejo2016ComputerAidedCO}. Noise is a dataset of square images where the resolution, contrast and pixel values are randomly generated (for details see Appendix \ref{images}). Finally, ImageNet-O \cite{Hendrycks2021NaturalAE} is a dataset OOD to ImageNet-1k that is adversarially constructed using a trained ResNet. Note that we exclude a number of OOD datasets from \cite{Kim2021AUB} and \cite{Huang2021MOSTS} as a result of discovering ID examples.

We train ResNet-50 \cite{He2016DeepRL}, DenseNet-121 \cite{Huang2017DenselyCC} and MobileNetV2 \cite{Sandler2018MobileNetV2IR} using hyperparameters based around standard ImageNet settings\footnote{\href{https://github.com/pytorch/examples/blob/main/imagenet/main.py}{https://github.com/pytorch/examples/blob/main/imagenet/main.py}}. Full training details can be found in Appendix \ref{training}. For each architecture we train 5 models independently using random seeds $\{1,\dots,5\}$ and report the mean result over the runs. Appendix \ref{results} additionally contains results on single pre-trained ImageNet-1k models, BiT ResNetV2-101 \cite{Kolesnikov2020BigT} and PyTorch DenseNet-121.

\subsubsection{Detection Methods for SCOD}
We consider four variations of SIRC using the components \{MSP,$\m H$\}$\times$\{$||\b z||_1,$Residual\}, as well as the components individually. We additionally evaluate various existing post-hoc methods: MSP \cite{Hendrycks2017ABF}, Energy \cite{Liu2020EnergybasedOD}, ViM \cite{Wang2022ViMOW} and Gradnorm \cite{Huang2021OnTI}. For SIRC and ViM we use the full ID train set to determine parameters.
Results for additional approaches, as well as further details pertaining to the methods, can be found in Appendix \ref{scores}. 

\subsection{Evaluation Metrics}\label{eval}
For evaluating different scoring functions $S$ for the SCOD problem setting we consider a number of metrics. Arrows($\uparrow\downarrow$) indicate whether higher/lower is better. (For graphical illustrations and additional metrics see Appendix \ref{metrics})
\begin{description}[style=unboxed,leftmargin=0cm]
\item[Area Under the Risk-Recall curve (AURR)$\downarrow$] We consider how empirical risk (Eq. \ref{risk}) varies with recall of ID\cmark, and aggregate performance over different $t$ by calculating the area under the curve. As recall is only measured over ID\cmark, the base accuracy of $f$ is not properly taken into account. Thus, this metric is only suitable for comparing different $g$ with $f$ fixed. To give an illustrative example, a $f,g$ pair where the classifier $f$ is only able to produce a single correct prediction will have perfect AURR as long as $S$ assigns that correct prediction the highest confidence (lowest uncertainty) score. Note that results for the AURC metric \cite{Kim2021AUB, Geifman2019BiasReducedUE} can be found in Appendix \ref{results}, although we omit them from the main paper as they are not notably different to AURR.

\item[Risk@Recall=0.95 (Risk@95)$\downarrow$] Since a rejection threshold $t$ must be selected at deployment, we also consider a particular setting of $t$ such that 95\% of ID\cmark is recalled. In practice, the corresponding value of $t$ could be found on a labelled ID validation set before deployment, without the use of any OOD data. It is worth noting that differences tend to be greater for this metric between different $S$ as it operates around the tail of the positive class.
\item[Area Under the ROC Curve (AUROC)$\uparrow$]
Since we are interested in rejecting both ID\xmark and OOD, we can consider ID\cmark as the positive class, and ID\xmark, OOD as separate negative classes. Then we can evaluate the AUROC of OOD|ID\cmark and ID\xmark|ID\cmark independently. The AUROC for a specific value of $\alpha$ would then be a weighted average of the two different AUROCs. This is not a direct measure of risk, but does measure the separation between different empirical distributions. Note that due to similar reasons to AURR this method is only valid for fixed $f$. 
\item[False Positive Rate@Recall=0.95 (FPR@95)$\downarrow$]
FPR@0.95 is similar to AUROC, but is taken at a specific $t$. It measures the proportion of the negative class accepted when the recall of the positive class (or true positive rate) is 0.95.

\end{description}

\subsection{Separation of ID\xmark|ID\cmark and OOD|ID\cmark Independently}\label{indep}

\begin{table}[t]
    \centering
\caption{\%AUROC and \%FPR@95 with ID\cmark as the positive class, considering ID\xmark and each OOD dataset separately. Full results are for ResNet-50 trained on ImageNet-200. We show abridged results for MobileNetV2 and DenseNet-121. \textbf{Bold} indicates best performance, \underline{underline} 2nd or 3rd best and we show the mean over models from 5 independent training runs. Variants of SIRC are shown as tuples of their components ($S_1$,$S_2$). We also show error rate on ID data. SIRC is able to consistently match or improve over $S_1$ for OOD|ID\cmark, at a negligible cost to ID\xmark|ID\cmark. Existing OOD detection methods are significantly worse for ID\xmark|ID\cmark and inconsistent at improving OOD|ID\cmark.}

\resizebox{\textwidth}{!}{
\begin{tabular}{lllbbaallllllll}

\toprule
 &  &  & \multicolumn{2}{c}{\textbf{ID\xmark}} & \multicolumn{2}{c}{\textbf{OOD mean}} & \multicolumn{2}{c}{\textbf{Near-IN-200}} & \multicolumn{2}{c}{\textbf{Caltech-45}} & \multicolumn{2}{c}{\textbf{Openimage-O}} & \multicolumn{2}{c}{\textbf{iNaturalist}} \\
\textbf{Model} &  & \textbf{Method}&AUROC$\uparrow$ & FPR@95$\downarrow$ & AUROC$\uparrow$ & FPR@95$\downarrow$ & AUROC$\uparrow$ & FPR@95$\downarrow$ & AUROC$\uparrow$ & FPR@95$\downarrow$ & AUROC$\uparrow$ & FPR@95$\downarrow$ & AUROC$\uparrow$ & FPR@95$\downarrow$ \\
\cmidrule(r){1-5} \cmidrule{6-15}
\multirow[c]{11}{*}{\begin{sideways}\shortstack[l]{\textbf{ResNet-50} \\ ID \%Error: 19.01}\end{sideways}} & \multirow[c]{4}{*}{\begin{sideways}\textbf{SIRC}\end{sideways}} & (MSP,$||\b z||_1$) & \underline{90.34} & \underline{52.70} & 91.51 & 40.27 & 85.56 & 59.76 & 91.36 & 41.44 & 92.28 & 41.36 & 94.80 & 29.60 \\
 &  & (MSP,Res.) & \textbf{90.43} & \textbf{52.10} & \underline{92.56} & \underline{34.98} & 85.52 & 60.03 & 91.19 & 42.27 & 92.57 & \underline{39.95} & 94.10 & 33.55 \\
 &  & ($-\mathcal{H}$,$||\b z||_1$) & 90.00 & 54.26 & 92.24 & 35.85 & \underline{85.88} & \underline{58.50} & \textbf{92.19} & \textbf{36.08} & \underline{92.87} & \underline{37.83} & \textbf{95.38} & \textbf{25.09} \\
 &  & ($-\mathcal{H}$,Res.) & 90.13 & 54.01 & \textbf{93.36} & \textbf{30.05} & \underline{85.85} & \underline{58.93} & \underline{92.11} & \underline{36.76} & \textbf{93.25} & \textbf{36.36} & \underline{94.82} & \underline{28.51} \\
 \cmidrule(r){2-5} \cmidrule{6-15}
 & \multirow[c]{7}{*}{} & MSP & \underline{90.41} & \underline{52.13} & 91.00 & 43.25 & 85.59 & 59.74 & 91.13 & 42.72 & 91.95 & 43.55 & 94.23 & 33.21 \\
 &  & $-\mathcal{H}$ & 90.07 & 54.05 & 91.81 & 38.24 & \textbf{85.91} & \textbf{58.47} & 92.01 & \underline{37.20} & \underline{92.59} & 40.10 & \underline{94.90} & \underline{28.01} \\
 &  & $||\b z||_1$ & 48.06 & 94.70 & 78.22 & 58.70 & 52.27 & 94.58 & 70.28 & 77.83 & 72.23 & 71.51 & 85.65 & 49.50 \\
 &  & Residual & 47.59 & 96.45 & 58.45 & 78.97 & 44.30 & 96.79 & 47.76 & 94.83 & 59.65 & 86.85 & 40.07 & 97.32 \\
 &  & Energy & 82.05 & 69.79 & 92.06 & \underline{35.32} & 81.96 & 68.70 & \underline{92.15} & 38.62 & 90.92 & 46.28 & 94.13 & 31.70 \\
 &  & Gradnorm & 60.17 & 87.88 & 85.22 & 44.41 & 62.90 & 86.89 & 81.11 & 59.23 & 81.09 & 57.80 & 91.00 & 34.46 \\
 &  & ViM & 80.62 & 78.13 & \underline{92.34} & 38.14 & 78.90 & 80.30 & 90.54 & 54.70 & 91.87 & 43.84 & 90.13 & 56.97 \\
\bottomrule
\end{tabular}
}
\resizebox{\textwidth}{!}{
\begin{tabular}{lllbbllllllllll}
\toprule
 &  &  & \multicolumn{2}{c}{\textbf{ID\xmark}} & \multicolumn{2}{c}{\textbf{Textures}} & \multicolumn{2}{c}{\textbf{Colonoscopy}} & \multicolumn{2}{c}{\textbf{Colorectal}} & \multicolumn{2}{c}{\textbf{Noise}} & \multicolumn{2}{c}{\textbf{ImageNet-O}} \\
\textbf{Model} &  & \textbf{Method} & AUROC$\uparrow$ & FPR@95$\downarrow$ & AUROC$\uparrow$ & FPR@95$\downarrow$ & AUROC$\uparrow$ & FPR@95$\downarrow$ & AUROC$\uparrow$ & FPR@95$\downarrow$ & AUROC$\uparrow$ & FPR@95$\downarrow$ & AUROC$\uparrow$ & FPR@95$\downarrow$ \\

\cmidrule(r){1-5} \cmidrule{6-15}

\multirow[c]{11}{*}{\begin{sideways}\shortstack[l]{\textbf{ResNet-50} \\ ID \%Error: 19.01}\end{sideways}} & \multirow[c]{4}{*}{\begin{sideways}\textbf{SIRC}\end{sideways}} & (MSP,$||\b z||_1$) & \underline{90.34} & \underline{52.70} & 93.64 & 32.02 & 95.93 & 25.33 & 95.84 & 24.39 & 90.72 & 49.63 & 83.44 & 58.91 \\
 &  & (MSP,Res.) & \textbf{90.43} & \textbf{52.10} & \underline{96.00} & \underline{19.81} & 95.52 & 27.31 & 95.32 & 26.97 & \underline{98.21} & \underline{10.97} & \underline{84.62} & \underline{53.99} \\
 &  & ($-\mathcal{H}$,$||\b z||_1$) & 90.00 & 54.26 & 94.38 & 27.38 & \underline{96.97} & \underline{16.87} & 96.71 & 18.71 & 91.74 & 45.84 & 84.01 & 56.34 \\
 &  & ($-\mathcal{H}$,Res.) & 90.13 & 54.01 & \underline{96.68} & \underline{15.70} & 96.72 & 18.10 & 96.41 & 20.42 & \underline{99.02} & \underline{4.89} & \underline{85.33} & \underline{50.81} \\
 
 \cmidrule(r){2-5} \cmidrule{6-15}
 
 & \multirow[c]{7}{*}{} & MSP & \underline{90.41} & \underline{52.13} & 92.88 & 36.61 & 95.75 & 26.52 & 94.86 & 30.28 & 89.33 & 56.83 & 83.29 & 59.78 \\
 &  & $-\mathcal{H}$ & 90.07 & 54.05 & 93.77 & 30.79 & \underline{96.87} & \underline{17.55} & 95.93 & 23.43 & 90.47 & 51.63 & 83.89 & 57.02 \\
 &  & $||\b z||_1$ & 48.06 & 94.70 & 88.90 & 39.67 & 76.97 & 82.24 & 97.28 & 14.64 & 97.36 & 13.51 & 63.00 & 84.82 \\
 &  & Residual & 47.59 & 96.45 & 82.84 & 46.63 & 38.09 & 99.64 & 53.93 & 88.78 & 91.31 & 20.92 & 68.04 & 78.98 \\
 &  & Energy & 82.05 & 69.79 & 95.37 & 22.50 & \textbf{97.51} & \textbf{14.19} & \textbf{99.07} & \underline{5.00} & 94.93 & 29.05 & 82.52 & 61.86 \\
 &  & Gradnorm & 60.17 & 87.88 & 93.00 & 26.57 & 90.54 & 42.85 & \underline{98.98} & \textbf{4.98} & 97.59 & 13.05 & 70.78 & 73.88 \\
 &  & ViM & 80.62 & 78.13 & \textbf{98.46} & \textbf{7.62} & 94.42 & 44.55 & \underline{98.04} & \underline{8.84} & \textbf{99.82} & \textbf{0.31} & \textbf{88.85} & \textbf{46.15} \\
\bottomrule
\end{tabular}
}
\resizebox{.85\textwidth}{!}{
\begin{tabular}{lllbbaa}
\toprule
 &  &  & \multicolumn{2}{c}{\textbf{ID\xmark}} & \multicolumn{2}{c}{\textbf{OOD mean}} \\
\textbf{Model} &  & \textbf{Method} & AUROC$\uparrow$ & FPR@95$\downarrow$ & AUROC$\uparrow$ & FPR@95$\downarrow$ \\
\cmidrule(r){1-5} \cmidrule{6-7}
\multirow[c]{11}{*}{\begin{sideways}\shortstack[l]{\textbf{MobileNetV2} \\ ID \%Error: 21.35}\end{sideways}} & \multirow[c]{4}{*}{\begin{sideways}\textbf{SIRC}\end{sideways}} & (MSP,$||\b z||_1$) & \underline{89.53} & \underline{55.51} & 92.27 & \underline{34.82} \\
 &  & (MSP,Res.) & \textbf{89.67} & \underline{55.10} & 91.78 & 38.56 \\
 &  & ($-\mathcal{H}$,$||\b z||_1$) & 88.90 & 58.64 & \textbf{92.92} & \textbf{32.16} \\
 &  & ($-\mathcal{H}$,Res.) & 89.12 & 57.85 & \underline{92.69} & \underline{34.20} \\
 
 \cmidrule(r){2-5} \cmidrule{6-7}
 & \multirow[c]{7}{*}{} & MSP & \underline{89.64} & \textbf{55.03} & 91.54 & 39.73 \\
 &  & $-\mathcal{H}$ & 89.02 & 58.43 & \underline{92.37} & 36.04 \\
 &  & $||\b z||_1$ & 53.56 & 93.40 & 81.06 & 53.50 \\
 &  & Residual & 41.99 & 97.30 & 41.42 & 94.11 \\
 &  & Energy & 81.87 & 67.98 & 91.68 & 36.68 \\
 &  & Gradnorm & 65.27 & 85.73 & 87.25 & 40.67 \\
 &  & ViM & 80.21 & 74.36 & 89.46 & 51.97 \\
\bottomrule
\end{tabular}\qquad 
\begin{tabular}{lllbbaa}
\toprule
 &  &  & \multicolumn{2}{c}{\textbf{ID\xmark}} & \multicolumn{2}{c}{\textbf{OOD mean}} \\
\textbf{Model} &  & \textbf{Method} & AUROC$\uparrow$ & FPR@95$\downarrow$ & AUROC$\uparrow$ & FPR@95$\downarrow$ \\
\cmidrule(r){1-5} \cmidrule{6-7}
\multirow[c]{11}{*}{\begin{sideways}\shortstack[l]{\textbf{DenseNet-121} \\ ID \%Error: 17.20}\end{sideways}} & \multirow[c]{4}{*}{\begin{sideways}\textbf{SIRC}\end{sideways}} & (MSP,$||\b z||_1$) & \underline{90.22} & \underline{52.41} & 91.68 & 38.83 \\
 &  & (MSP,Res.) & \underline{90.20} & \underline{52.42} & \underline{92.81} & \underline{32.68} \\
 &  & ($-\mathcal{H}$,$||\b z||_1$) & 89.95 & 53.96 & \underline{92.42} & \underline{32.92} \\
 &  & ($-\mathcal{H}$,Res.) & 89.92 & 54.17 & \textbf{93.45} & \textbf{27.97} \\
 \cmidrule(r){2-5} \cmidrule{6-7}
 
 & \multirow[c]{7}{*}{} & MSP & \textbf{90.30} & \textbf{51.85} & 91.44 & 40.44 \\
 &  & $-\mathcal{H}$ & 90.04 & 53.41 & 92.24 & 34.49 \\
 &  & $||\b z||_1$ & 36.87 & 98.70 & 63.53 & 80.35 \\
 &  & Residual & 46.08 & 95.44 & 69.38 & 71.33 \\
 &  & Energy & 82.12 & 66.54 & 90.92 & 38.87 \\
 &  & Gradnorm & 50.18 & 95.19 & 76.18 & 62.58 \\
 &  & ViM & 76.63 & 84.73 & 90.50 & 44.71 \\
\bottomrule
\end{tabular}
}
    \label{tab:auroc}
\end{table}

Table \ref{tab:auroc} shows \%AUROC and \%FPR@0.95 with ID\cmark as the positive class and ID\xmark, OOD independently as different negative classes (see Section \ref{eval}). In general, we see that SIRC, compared to $S_1$, is able to improve OOD|ID\cmark whilst incurring only a small ($<0.2$\%AUROC) reduction in the ability to distinguish ID\xmark|ID\cmark, across all 3 architectures. On the other hand, non-softmax methods designed for OOD detection show poor ability to identify ID\xmark, with performance ranging from $\sim 8$ worse \%AUROC than MSP to $\sim 50\%$ AUROC (random guessing). Furthermore, they cannot consistently outperform the baseline when separating OOD|ID\cmark, in line with the discussion in Section \ref{OODbad}.
\subsubsection{SIRC is Robust to Weak $S_2$}
Although for the majority of OOD datasets SIRC is able to outperform $S_1$, this is not always the case. For these latter instances, we can see that $S_2$ individually is not useful, e.g. for ResNet-50 on Colonoscopy, Residual performs \textit{worse} than random guessing. However, in cases like this the performance is still close to that of $S_1$. As $S_2$ will tend to be higher for these OOD datasets, the behaviour is like that for ID\xmark|ID, with the decision boundaries close to vertical (see Fig. \ref{fig:combs}). As such SIRC is \textit{robust} to $S_2$ performing poorly, but is able to improve on $S_1$ when $S_2$ is of use. In comparison, ViM, which linearly combines Energy and Residual, is much more sensitive to when the latter stumbles. On Colonoscopy ViM has $\sim 30$ worse \%FPR@95 compared to Energy, whereas SIRC ($-\m H$, Res.) loses $<1\%$ compared to $-\m H$.

\subsubsection{OOD Detection Methods are Inconsistent Over Different Data}
The performance of existing methods for OOD detection relative to the MSP baseline is varies considerably from dataset to dataset. For example, even though ViM is able to perform very well on Textures, Noise and ImageNet-O (>50 better \%FPR@95 on Noise), it does worse than the baseline on most other OOD datasets (>20 worse \%FPR@95 for Near-ImageNet-200 and iNaturalist). This suggests that the inductive biases incorporated, and assumptions made, when designing existing OOD detection methods may prevent them from generalising across a wider variety of OOD data. In contrast, SIRC more \textit{consistently}, albeit modestly, improves over the baseline, due to its aforementioned robustness.

\subsection{Varying the Importance of OOD Data Through $\alpha$ and $\beta$}\label{vary}
\begin{figure}[t]
    \centering
    \includegraphics[width=\linewidth]{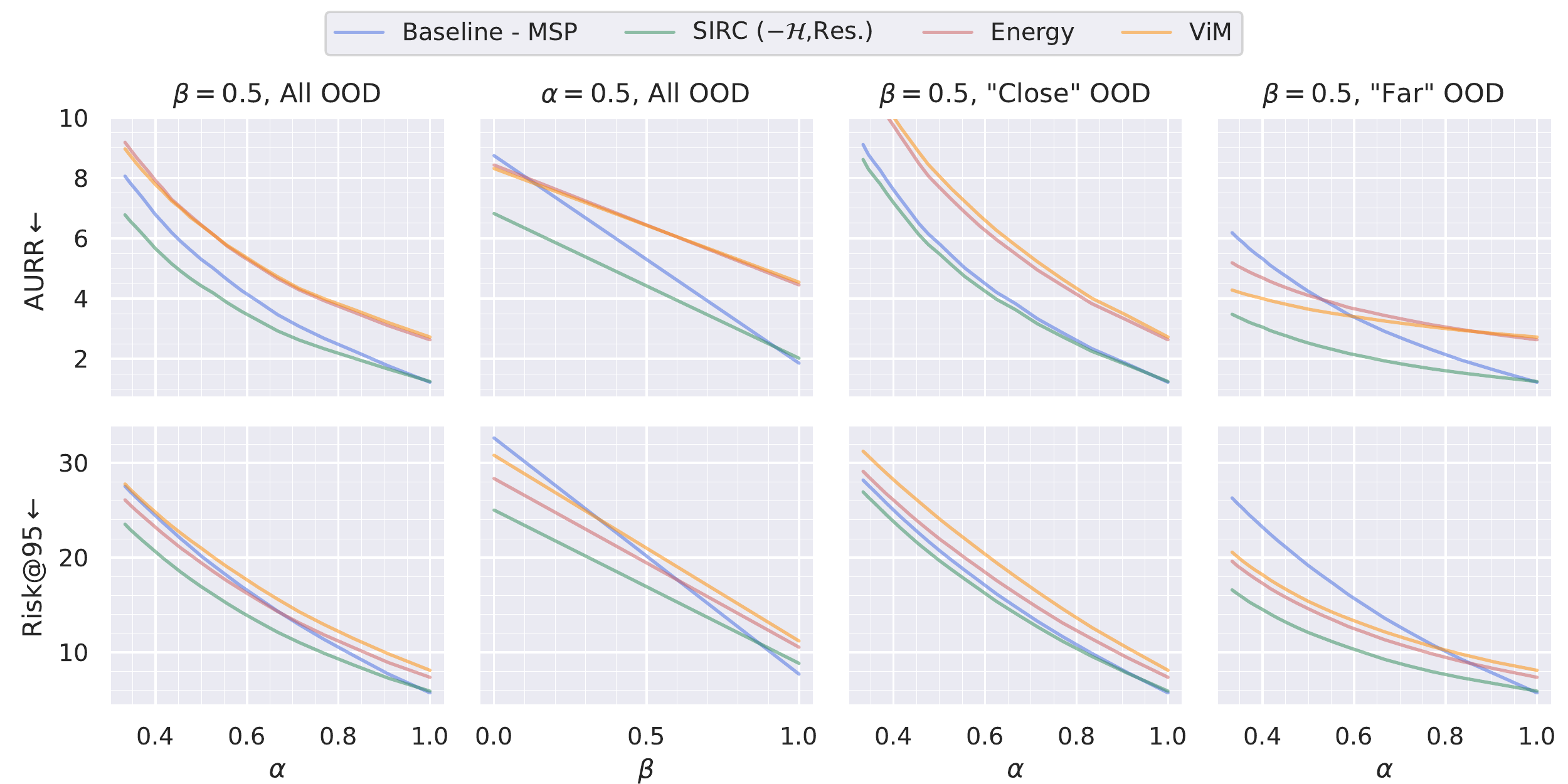}
    \caption{AURR$\downarrow$ and Risk@95$\downarrow$ ($\times 10^2$) for different methods as $\alpha$ and $\beta$ vary (Eqs. \ref{mix},\ref{loss}) on a mixture of all the OOD data. We also split the OOD data into qualitatively ``Close'' and ``Far'' subsets (Section \ref{vary}). For high $\alpha, \beta$, where ID\xmark dominates in the risk, the MSP \textcolor{RoyalBlue}{baseline} is the best. As $\alpha, \beta$ decrease, increasing the effect of OOD data, other methods improve relative to the \textcolor{RoyalBlue}{baseline}. \textcolor{ForestGreen}{SIRC} is able to \textit{most consistently} improve over the \textcolor{RoyalBlue}{baseline}. OOD detection methods perform better on ``Far'' OOD. The ID dataset is ImageNet-200, the model ResNet-50. We show the mean over 5 independent training runs. We multiply all values by $10^2$ for readability.}
    \label{fig:vary}
\end{figure}
At deployment, there will be a specific ratio of ID:OOD data exposed to the model. Thus, it is of interest to investigate the risk over different values of $\alpha$ (Eq. \ref{mix}). Similarly, an incorrect ID prediction may or may not be more costly than a prediction on OOD data so we investigate different values of $\beta$ (Eq. \ref{loss}). Fig. \ref{fig:vary} shows how AURR and Risk@95 are affected as $\alpha$ and $\beta$ are varied independently (with the other fixed to 0.5). We use the full test set of ImageNet-200, and pool OOD datasets together and sample different quantities of data randomly in order to achieve different values of $\alpha$. We use 3 different groupings of OOD data: All, ``Close'' \{Near-ImageNet-200, Caltech-45, Openimage-O, iNaturalist\} and ``Far'' \{Textures, Colonoscopy, Colorectal, Noise\}. These groupings are based on relative qualitative semantic difference to the ID dataset (see Appendix \ref{images} for example images from each dataset). Although the grouping is not formal, it serves to illustrate OOD data-dependent differences in SCOD performance. 
\subsubsection{Relative Performance of Methods Changes with $\alpha$ and $\beta$}
At high $\alpha$ and $\beta$, where ID\xmark dominates the risk, the MSP baseline performs best. However, as $\alpha$ and $\beta$ are decreased, and OOD data is introduced, we see that other methods improve relative to the baseline. There may be a \textit{crossover} after which the ability to better distinguish OOD|ID\cmark allows a method to surpass the baseline. Thus, which method to choose for deployment will depend on the practitioner's setting of $\beta$ and (if they have any knowledge of it at all) of $\alpha$.
\subsubsection{SIRC Most Consistently Improves Over the Baseline}
 SIRC $(-\m H, \text{Res.})$ is able to outperform the baseline most consistently over the different scenarios and settings of $\alpha, \beta$, only doing worse for ID\xmark dominated cases ($\alpha, \beta$ close to 1). This is because SIRC has close to baseline ID\xmark|ID\cmark performance and is superior for OOD|ID\cmark. In comparison, ViM and Energy, which conflate ID\xmark and ID\cmark, are often worse than the baseline for most (if not all) values of $\alpha, \beta$. Their behaviour on the different groupings of data illustrates how these methods may be biased towards different OOD datasets, as they significantly outperform the baseline at lower $\alpha$ for the ``Far'' grouping, but always do worse on ``Close'' OOD data. 

\subsection{Comparison Between SCOD and OOD Detection}\label{sc_comp}
\begin{figure}[t]
    \centering
    \includegraphics[width=\linewidth]{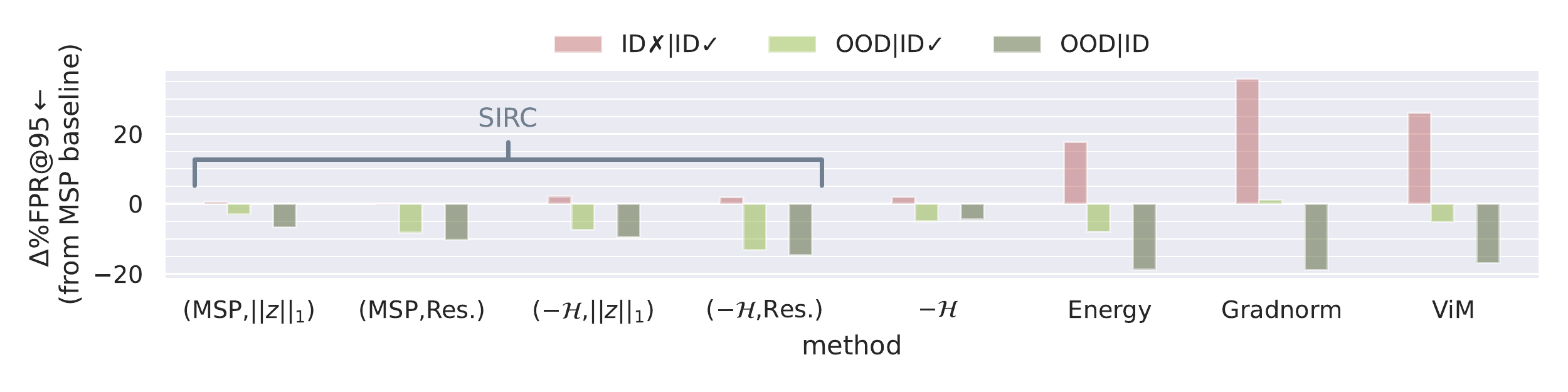}
    \caption{The change in \%FPR@95$\downarrow$ relative to the MSP baseline of different methods. Different data classes are shown negative|positive. Although OOD detection methods are able to improve \textcolor{OliveGreen}{OOD|ID}, they do so mainly at the expense of \textcolor{BrickRed}{ID\xmark|ID\cmark} rather than improving \textcolor{LimeGreen}{OOD|ID\cmark}. SIRC is able to improve \textcolor{LimeGreen}{OOD|ID\cmark} with minimal loss to \textcolor{BrickRed}{ID\xmark|ID\cmark}, alongside modest improvements for \textcolor{OliveGreen}{OOD|ID}. Results for OOD are averaged over all OOD datasets. The ID dataset is ImageNet-200 and the model ResNet-50.}
    \label{fig:ood_scod}
\end{figure}
Fig. \ref{fig:ood_scod} shows the difference in \%FPR@95 relative to the MSP baseline for different combinations of negative|positive data classes (ID\xmark|ID\cmark, OOD|ID\cmark, OOD|ID), where OOD results are averaged over all datasets and training runs. In line with the discussion in Section \ref{OODbad}, we observe that the non-softmax OOD detection methods are able to improve over the baseline for OOD|ID, but this comes mostly at the cost of inferior ID\xmark|ID\cmark rather than due to better OOD|ID\cmark, so they will do worse for SCOD. SIRC on the other hand is able to retain much more ID\xmark|ID\cmark performance whilst improving on OOD|ID\cmark, allowing it to have better OOD detection \textit{and} SCOD performance compared to the baseline.

\section{Related Work}
There is extensive existing research into OOD detection, a survey of which can be found in \cite{Yang2021GeneralizedOD}. To improve over the MSP baseline in \cite{Hendrycks2017ABF}, early post-hoc approaches, primarily experimenting on CIFAR-scale data, such as ODIN \cite{Liang2018EnhancingTR}, Mahalanobis \cite{Lee2018ASU}, Energy \cite{Liu2020EnergybasedOD} explore how to extract non-softmax information from a trained network. 
 More recent work has moved to larger-scale image datasets \cite{Huang2021MOSTS, Hendrycks2020ScalingOD}. Gradnorm \cite{Huang2021OnTI}, although motivated by the information in gradients, at its core combines information from the softmax and features together. Similarly, ViM \cite{Wang2022ViMOW} combines Energy with the class-agnostic Residual score. ReAct \cite{Sun2021ReActOD} aims to improve logit/softmax-based scores by clamping the magnitude of final layer features. There are also many training-based approaches. Outlier Exposure \cite{Hendrycks2019DeepAD} explores training networks to be uncertain on ``known'' existing OOD data, whilst VOS \cite{Du2022VOSLW} instead generates virtual outliers during training for this purpose.
 \cite{Hsu2020GeneralizedOD, Techapanurak_2020_ACCV} propose the network explicitly learn a scaling factor for the logits to improve softmax behaviour. There also exists a line of research that explores the use of generative models, $p(\b x;\b \theta)$, for OOD detection \cite{Caterini2021EntropicII,Zhang2021OnTO, likelihoodratio, Nalisnick2019DoDG}, however, these approaches are completely separate from classification.

Selective classification, or misclassification detection, has also been investigated for deep learning scenarios.
Initially examined in \cite{Geifman2017SelectiveCF, Hendrycks2017ABF}, there are a number of approaches to the task that target the classifier $f$ through novel training losses and/or architectural adjustments  \cite{Moon2020ConfidenceAwareLF,failure, geifman2019selectivenet}. Post-hoc approaches are fewer. DOCTOR \cite{Granese2021DOCTORAS} provides theoretical justification for using the $l_2$-norm of the softmax output $||\b \pi||_2$ as a confidence score for detecting misclassifications, however, we find its behaviour similar to MSP and $\m H$  (See Appendix \ref{results}).

There also exist general approaches for uncertainty estimation that are then evaluated using the above tasks, e.g. Bayesian Neural Networks \cite{bnn}, MC-Dropout \cite{pmlr-v48-gal16}, Deep Ensembles \cite{Lakshminarayanan2017SimpleAS}, Dirichlet Networks \cite{Malinin2018PredictiveUE, Malinin2020EnsembleDD} and DDU \cite{Mukhoti2021DeterministicNN}.

The two works closest to ours are \cite{Kamath2020SelectiveQA} and \cite{Kim2021AUB}. \cite{Kamath2020SelectiveQA} investigates selective classification under covariate shift for the natural language processing task of question and answering. In the case of \textit{covariate} shift, valid predictions can still be produced on the shifted data, which by our definition is not possible for OOD data (see Section \ref{prelim}). Thus the problem setting here is different to our work. 
We remark that it would be of interest to extend this work to investigate selective classification with covariate shift for tasks in computer vision. \cite{Kim2021AUB} introduces the idea that ID\xmark and OOD data should be rejected together and investigates the performance of a range of existing approaches. They examine both training and post-hoc methods (comparing different $f$ and $g$) on SCOD (which they term unknown detection), as well as misclassification detection and OOD detection. They do not provide a novel approach targeting SCOD, and consider a single setting of ($\alpha, \beta$), where the $\alpha$ is not specified and $\beta = 0.5$.
\section{Concluding Remarks}
In this work, we consider the performance of existing methods for OOD detection on selective classification in the presence of out-of-distribution data (SCOD). We show how their improved OOD detection vs the MSP baseline often comes at the cost of inferior SCOD performance. Furthermore, we find their performance is inconsistent over different OOD datasets. In order to improve SCOD performance over the baseline, we develop SIRC. Our approach aims to retain information, which is useful for detecting misclassifications, from a softmax-based confidence score, whilst incorporating additional information useful for identifying OOD samples. Experiments show that SIRC is able to consistently match or improve over the baseline approach for a wide range of datasets, CNN architectures and problem scenarios. We hope this work encourages the further investigation of SCOD or that of other new detection tasks. 

\renewcommand\bibname{References} 
\bibliographystyle{splncs_natbib}
\bibliography{bib}
\newpage
\appendix
\section{Experimental Details}
We present detailed information about our experimental setup. Our code is available at \url{https://github.com/Guoxoug/SIRC}.
\subsection{Models and Training}\label{training}
For the main results we train ResNet-50 \cite{He2016DeepRL} using the default hyperparameters found in PyTorch's examples.\footnote{\url{https://github.com/pytorch/examples/tree/main/imagenet}} We train on ImageNet-200 for 90 epochs with a batch size of 256. Stochastic gradient descent is used with a weight decay of $10^{-4}$, a momentum of 0.9 and an initial learning rate of 0.1 that steps down by a factor of 10 at epochs 30 and 60. Images are augmented using \texttt{RandomResizedCrop} and \texttt{RandomHorizontalFlip}. MobileNetV2 \cite{Sandler2018MobileNetV2IR} uses the same setting, but with an initial learning rate of 0.05. DenseNet-121 is trained with the same settings are ResNet-50 but with Nesterov momentum as per \cite{Huang2017DenselyCC}. We perform 5 independent training runs for each architecture, with random seeds $\{1,...,5\}$.

Additionally, we also test on two pre-trained ImageNet-1k models. We use ResNetV2-101 from Google's Big Transfer\footnote{\url{https://github.com/google-research/big_transfer}} \cite{Kolesnikov2020BigT}, specifically \texttt{BiT-S-R101x1}, and DenseNet-121 provided by PyTorch.\footnote{\url{https://pytorch.org/vision/stable/models.html}}. Note that the BiT model takes $480\times480$ images as input, whereas all other models take standard ImageNet-scale $224\times224$ images. Note that for evaluating these models we exclude Near-ImageNet-200 and Caltech-45 due to class overlap with ImageNet-1k.
\subsection{ImageNet-Scale Datasets}\label{images}
\begin{figure}
    \centering
    \includegraphics[width=\linewidth]{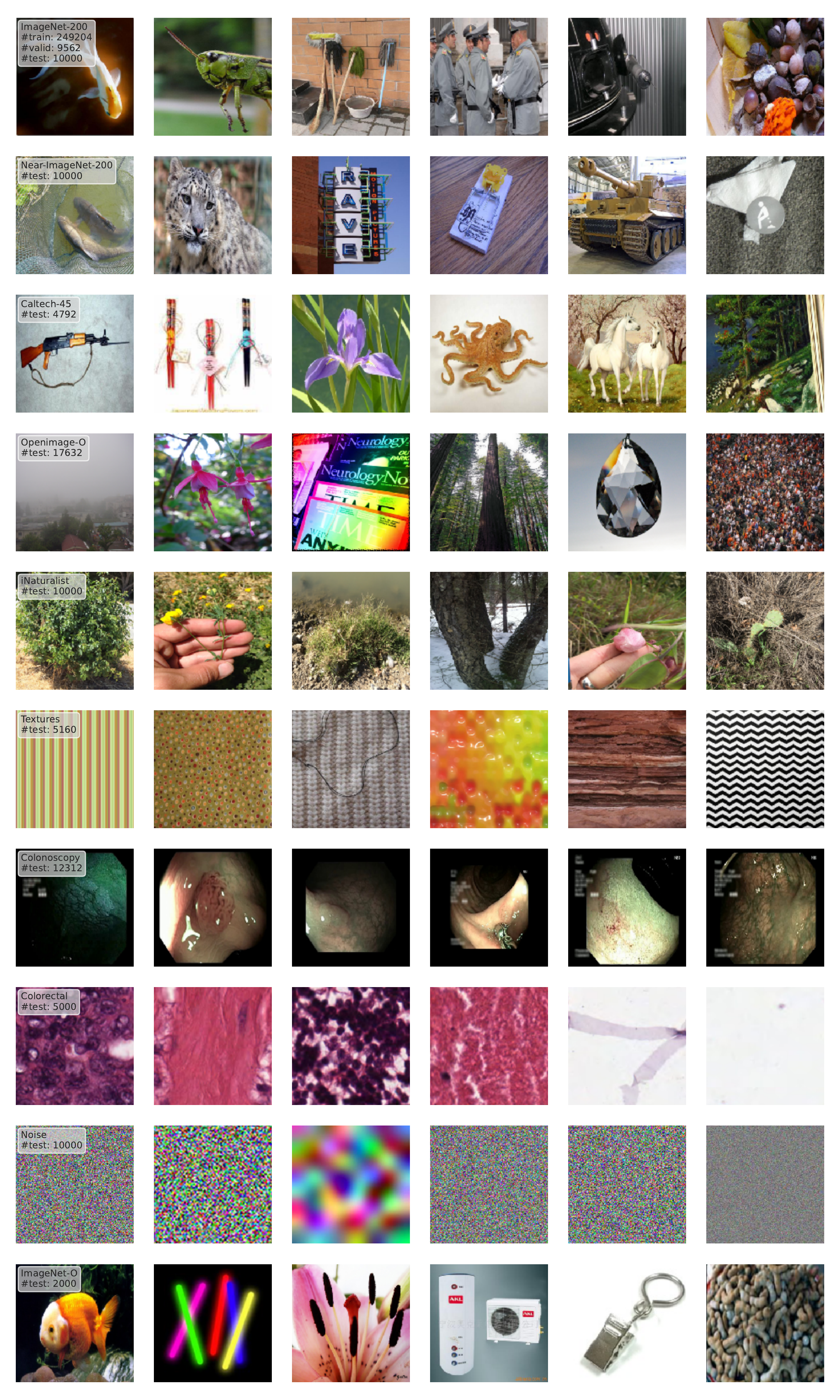}
    \caption{Random examples from each ImageNet-scale dataset, with the \#samples in each.}
    \label{fig:images}
\end{figure}

Figure \ref{fig:images} shows a number of random examples from each dataset introduced in Section \ref{exp}, alongside the number of samples in said dataset. Below we describe the methodology for constructing Colonoscopy and Noise. For the remaining datasets please refer to their original papers for details \cite{Huang2021MOSTS, Wang2022ViMOW, Kim2021AUB, Hendrycks2021NaturalAE, Kather2016MulticlassTA}. We note that there is a slight discrepancy between the number of samples reported in \cite{Kim2021AUB} for ImageNet-200 and in the authors' provided datasets,\footnote{\url{https://github.com/daintlab/unknown-detection-benchmarks}} but we do not believe this affects the validity of our results. 
\subsubsection{Noise}
We randomly generate 10000 square images. All samples are generated independently. Within each image, each value (in space and RGB) is sampled from the same gaussian distribution, with mean 0.5. The standard deviation of said gaussian differs between images. These in turn are generated by sampling from a unit gaussian and squaring the samples. Pixel values are then clipped to be in $[0,1]$ and mapped to 8-bit integers. The widths of each image are sampled uniformly from $\{2,...,256\}$, and the images are all scaled to $256\times256$ using the lanczos interpolation method in PIL.\footnote{\url{https://pillow.readthedocs.io/en/stable/_modules/PIL/Image.html\#Image.resize}} The resulting data thus varies in both scale and contrast (see Fig. \ref{fig:images}).
\subsubsection{Colonoscopy}
We separate out frames as individual images from videos provided in \cite{Mesejo2016ComputerAidedCO}.\footnote{\url{http://www.depeca.uah.es/colonoscopy_dataset/}} We download the first 10 narrow band imaging (NBI) videos in each class of lesion (hyperplasic, serrated, adenoma) and extract each frame as an individual image. Although the data is not independent in this case, we treat it as such for the purposes of our investigation.

\subsection{Confidence Scores}\label{scores}
Below we detail all confidence scores $S$ implemented and evaluated in our investigation. There are additional approaches that were omitted from the main paper for the sake of brevity.
\begin{itemize}
    \item SIRC: for a description of the score see Section \ref{sirc} in the main paper. We use the whole of the ImageNet-200 \textit{training} set to determine the values of $\mu_{S_2},\sigma_{S_2}$. For ImageNet-1k we randomly sample 250,000 images from the training set. Note that for all following methods that require ID data to find parameters, we use the same ID data as for SIRC. We investigate combinations of $S_1,S_2$ from the cartesian product \{MSP,DOCTOR,$\m H$\}$\times$\{$||\b z||_1,$Residual\}.
    \item Maximum Softmax Probability (MSP)\cite{Hendrycks2017ABF}: a baseline score that takes the max value from the softmax $\pi_\text{max} = \max_k \pi_k$.
    \item DOCTOR \cite{Granese2021DOCTORAS}: the original paper does not directly present it as such, but the confidence score is equivalent to $||\b \pi||_2$.
    \item Softmax Entropy ($\m H$): measures softmax uncertainty, $\m H[\b \pi] = -\sum_k \pi_k\log\pi_k$. We use $S = -\m H[\b \pi]$ to change it to a measure of confidence. 
    \item $l_1$-norm of the features: used in Gradnorm \cite{Huang2021OnTI}, $||\b z||_1$. 
    \item Residual: used in ViM \cite{Huang2021OnTI}, this score measures the component of the feature vector that is outside of a principal subspace defined using ID data, $||\b z^{P^\bot}||_2$. We follow \cite{Wang2022ViMOW} in setting the dimensionality of the subspace to 1000 if the dimensionality of $\b z$, $L>1500$ and 512 otherwise. Like Entropy, we use the negative of the score $S = -||\b z^{P^\bot}||_2$ as this score is meant to be higher for OOD data. Please refer to \citet{Wang2022ViMOW}'s paper for full details.
    \item Max Logit \cite{Hendrycks2020ScalingOD}: Max Logit is similar to MSP, but the score is taken from the logits before the softmax $v_\text{max} = \max_k v_k$.
    \item Energy \cite{Liu2020EnergybasedOD}: this score aggregates over all logit values as $\log \sum_k \exp v_k$.
    \item Gradnorm \cite{Huang2021OnTI}: although this score was originally motivated by gradients, we can view it simply as the combination of two scores, $C = ||\b \pi - \b 1/K||_1||\b z||_1$.
    \item ViM \cite{Wang2022ViMOW}: this linearly combines Energy and Residual, $C = \log \sum_k \exp v_k - c||\b z^{P^\bot}||_2$. The parameter $c$ is given by the average value of Max Logit divided by the average value of Residual on ID data, which scales the importance of Residual to be similar to that of Energy in the combination. 
    \item Mahalanobis \cite{Lee2018ASU}: this score involves building a classwise gaussian mixture model over the features with tied covariance matrix. The confidence is then calculated as $ -\min_k (\b z - \b \mu_k)^T  \Tilde{\b \Sigma} (\b z - \b \mu_k)$. We use the approach in \cite{Wang2022ViMOW, Fort2021ExploringTL} where only the final layer features are considered.
\end{itemize}
\subsection{Evaluation Metrics}\label{metrics}
\begin{figure}[t]
    \centering
    \includegraphics[width=\linewidth]{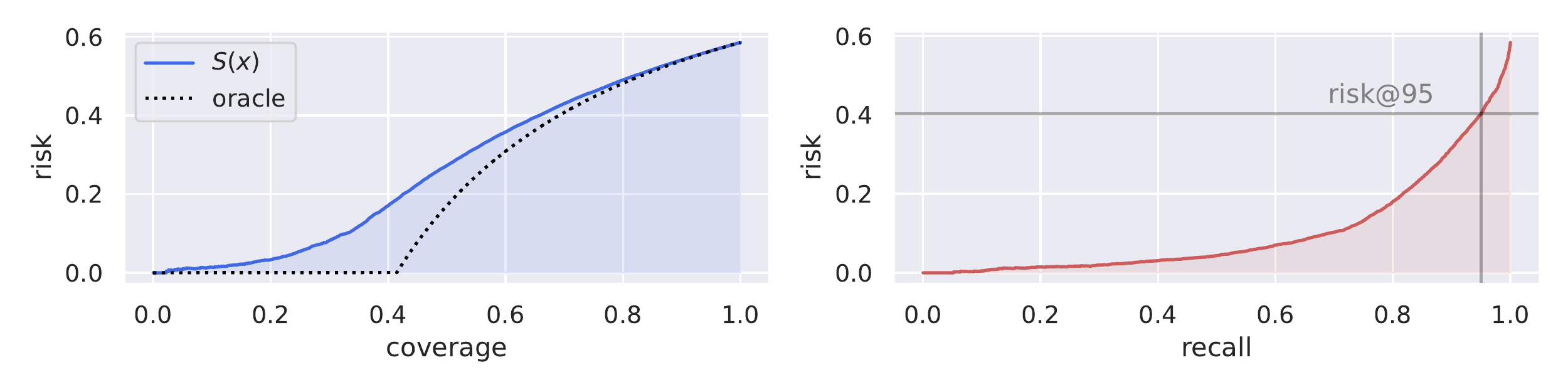}
    \caption{Visualisations of different evaluation metrics for SCOD. We aim to minimise risk over different selection thresholds $t$. \textcolor{NavyBlue}{Left}: Risk-Coverage curve (coverage is the proportion of all data accepted). We aggregate performance over $t$ by taking the area under the curve. The oracle represents perfect separation of OOD,ID\xmark|ID\cmark. \textcolor{BrickRed}{Right}: Risk-Recall curve. We consider both the area under the curve as well as risk@recall=0.95.}
    \label{fig:metrics}
\end{figure}

Other than the metrics specified in Section \ref{eval}, we additionally use Area Under the Risk-Coverage Curve (AURC)$\downarrow$, from \cite{Kim2021AUB, Geifman2017SelectiveCF}. It aggregates risk over all values of \textit{coverage}, which is the proportion of all input data accepted. For AURC their exists an oracle curve, where OOD and ID\xmark are perfectly disjoint from ID\cmark. AURC can be reduced either by lowering the oracle curve by reducing the number of ID\xmark (increasing baseline accuracy of $f$) or by better separating OOD,ID\xmark|ID\cmark (better choice of $g$) and so bringing the curve closer to the oracle. Thus the metric is suitable for both training based, and post-hoc approaches. Fig. \ref{fig:metrics} illustrates graphically some of the metrics we use to evaluate SCOD.
\section{Additional Results}\label{results}
We provide more complete versions of the results presented in Section \ref{exp} of the main work across all architectures and datasets. 

\subsection{AUROC and FPR@95}
We present results across all post-hoc confidence scores in Appendix \ref{scores} for all architectures. We also include mean$\pm 2$std. for experiments with multiple training runs. SIRC performs as expected in all cases -- a negligible reduction in ID\xmark|ID\cmark in exchange for a meaningful uplift in OOD|ID\cmark compared to only using $S_1$.
DOCTOR in general performs somewhere in between MSP and $-\m H$, both individually and when used in SIRC, so we relegate it to the appendix. We note that Residual and Mahalanobis perform much better only for ResNetV2-101 (these results are inline with \cite{Wang2022ViMOW}). This may be due to the fact that BiT uses Weight Standardisation and Group Normalisation when training, rather than standard Batch Normalisation. \citet{Mukhoti2021DeterministicNN} show that limiting the Lipschitz constant of the network during training improves the OOD detection performance of gaussian mixture models, which may be also what is occurring in this example. The Mahalanobis detector performs poorly outside ResNetV2-101 otherwise. There is non-negligible variance between training runs on a number of OOD datasets, highlighting the need to perform multiple training runs. Some datasets (e.g. Noise,  Colorectal), have especially high variation.

\begin{table}[]
    \centering
        \caption{Full \%AUROC and \%FPR@95 results for all models trained on ImageNet-200. We show the mean$\pm2$std. over 5 independent training runs.  \textbf{Bold} indicates best performance, \underline{underline} 2nd or 3rd best.}
    \label{tab:imagenet200_full}

\resizebox{.85\textwidth}{!}{
\begin{tabular}{lllbbaallllllll}

\toprule
 &  &  & \multicolumn{2}{c}{\textbf{ID\xmark}} & \multicolumn{2}{c}{\textbf{OOD mean}} & \multicolumn{2}{c}{\textbf{Near-IN-200}} & \multicolumn{2}{c}{\textbf{Caltech-45}} & \multicolumn{2}{c}{\textbf{Openimage-O}} & \multicolumn{2}{c}{\textbf{iNaturalist}} \\
\textbf{Model} &  & \textbf{Method}&\%AUROC$\uparrow$ & \%FPR@95$\downarrow$ & \%AUROC$\uparrow$ & \%FPR@95$\downarrow$ & \%AUROC$\uparrow$ & \%FPR@95$\downarrow$ & \%AUROC$\uparrow$ & \%FPR@95$\downarrow$ & \%AUROC$\uparrow$ & \%FPR@95$\downarrow$ & \%AUROC$\uparrow$ & \%FPR@95$\downarrow$ \\
\cmidrule(r){1-5} \cmidrule{6-15}

\multirow[c]{16}{*}{\begin{sideways}\shortstack[l]{\textbf{ResNet-50} \\ ID \%Error: 19.01}\end{sideways}} & \multirow[c]{6}{*}{\begin{sideways}\textbf{SIRC}\end{sideways}} & (MSP,$||\b z||_1$) & 90.34 \scriptsize ±0.2 & 52.70 \scriptsize ±3.2 & 91.51 & 40.27 & 85.56 \scriptsize ±0.6 & 59.76 \scriptsize ±2.9 & 91.36 \scriptsize ±0.6 & 41.44 \scriptsize ±3.0 & 92.28 \scriptsize ±0.5 & 41.36 \scriptsize ±2.8 & 94.80 \scriptsize ±0.3 & 29.60 \scriptsize ±1.3 \\
 &  & (MSP,Res.) & \textbf{90.43 \scriptsize ±0.3} & \underline{52.10 \scriptsize ±3.0} & \underline{92.56} & 34.98 & 85.52 \scriptsize ±0.6 & 60.03 \scriptsize ±2.4 & 91.19 \scriptsize ±0.6 & 42.27 \scriptsize ±3.2 & 92.57 \scriptsize ±0.6 & 39.95 \scriptsize ±3.3 & 94.10 \scriptsize ±0.3 & 33.55 \scriptsize ±2.7 \\
 &  & (DR,$||\b z||_1$) & 90.29 \scriptsize ±0.3 & 52.54 \scriptsize ±3.4 & 91.83 & 37.08 & 85.68 \scriptsize ±0.6 & \underline{58.05 \scriptsize ±2.9} & 91.67 \scriptsize ±0.5 & 37.81 \scriptsize ±2.7 & 92.59 \scriptsize ±0.4 & \underline{37.82 \scriptsize ±2.6} & \underline{95.18 \scriptsize ±0.3} & \underline{25.78 \scriptsize ±1.7} \\
 &  & (DR,Res.) & \underline{90.40 \scriptsize ±0.4} & \textbf{51.81 \scriptsize ±2.9} & \underline{92.83} & \underline{31.76} & 85.62 \scriptsize ±0.6 & 58.49 \scriptsize ±3.2 & 91.44 \scriptsize ±0.6 & 38.92 \scriptsize ±3.2 & \underline{92.87 \scriptsize ±0.6} & \textbf{36.36 \scriptsize ±3.6} & 94.32 \scriptsize ±0.4 & 30.05 \scriptsize ±3.2 \\
 &  & ($-\mathcal{H}$,$||\b z||_1$) & 90.00 \scriptsize ±0.4 & 54.26 \scriptsize ±2.7 & 92.24 & 35.85 & \underline{85.88 \scriptsize ±0.6} & 58.50 \scriptsize ±3.3 & \underline{92.19 \scriptsize ±0.5} & \textbf{36.08 \scriptsize ±2.5} & \underline{92.87 \scriptsize ±0.4} & 37.83 \scriptsize ±3.3 & \textbf{95.38 \scriptsize ±0.2} & \textbf{25.09 \scriptsize ±1.9} \\
 &  & ($-\mathcal{H}$,Res.) & 90.13 \scriptsize ±0.4 & 54.01 \scriptsize ±3.2 & \textbf{93.36} & \textbf{30.05} & \underline{85.85 \scriptsize ±0.6} & 58.93 \scriptsize ±3.3 & 92.11 \scriptsize ±0.5 & \underline{36.76 \scriptsize ±3.0} & \textbf{93.25 \scriptsize ±0.6} & \underline{36.36 \scriptsize ±4.5} & 94.82 \scriptsize ±0.3 & 28.51 \scriptsize ±3.4 \\
 \cmidrule(r){2-5} \cmidrule{6-15}

& \multirow[c]{10}{*}{} & MSP & \underline{90.41 \scriptsize ±0.3} & 52.13 \scriptsize ±2.0 & 91.00 & 43.25 & 85.59 \scriptsize ±0.6 & 59.74 \scriptsize ±2.0 & 91.13 \scriptsize ±0.6 & 42.72 \scriptsize ±2.8 & 91.95 \scriptsize ±0.5 & 43.55 \scriptsize ±2.4 & 94.23 \scriptsize ±0.3 & 33.21 \scriptsize ±1.2 \\
 &  & DOCTOR & 90.39 \scriptsize ±0.3 & \underline{51.87 \scriptsize ±1.6} & 91.26 & 40.22 & 85.73 \scriptsize ±0.6 & \textbf{57.89 \scriptsize ±2.3} & 91.41 \scriptsize ±0.5 & 39.22 \scriptsize ±2.2 & 92.20 \scriptsize ±0.5 & 40.22 \scriptsize ±2.7 & 94.51 \scriptsize ±0.3 & 29.41 \scriptsize ±1.8 \\
 &  & $-\mathcal{H}$ & 90.07 \scriptsize ±0.4 & 54.05 \scriptsize ±2.9 & 91.81 & 38.24 & \textbf{85.91 \scriptsize ±0.6} & \underline{58.47 \scriptsize ±3.3} & 92.01 \scriptsize ±0.5 & 37.20 \scriptsize ±2.6 & 92.59 \scriptsize ±0.5 & 40.10 \scriptsize ±3.9 & \underline{94.90 \scriptsize ±0.3} & \underline{28.01 \scriptsize ±3.2} \\
 &  & $||\b z||_1$ & 48.06 \scriptsize ±1.1 & 94.70 \scriptsize ±1.4 & 78.22 & 58.70 & 52.27 \scriptsize ±0.7 & 94.58 \scriptsize ±0.5 & 70.28 \scriptsize ±1.6 & 77.83 \scriptsize ±1.8 & 72.23 \scriptsize ±2.4 & 71.51 \scriptsize ±2.6 & 85.65 \scriptsize ±2.7 & 49.50 \scriptsize ±5.8 \\
 &  & Residual & 47.59 \scriptsize ±1.8 & 96.45 \scriptsize ±1.1 & 58.45 & 78.97 & 44.30 \scriptsize ±1.1 & 96.79 \scriptsize ±0.4 & 47.76 \scriptsize ±1.4 & 94.83 \scriptsize ±0.9 & 59.65 \scriptsize ±4.0 & 86.85 \scriptsize ±2.2 & 40.07 \scriptsize ±6.3 & 97.32 \scriptsize ±0.7 \\
 &  & Max Logit & 83.21 \scriptsize ±0.6 & 65.16 \scriptsize ±3.4 & 92.33 & \underline{34.15} & 82.68 \scriptsize ±0.7 & 65.37 \scriptsize ±3.6 & \textbf{92.48 \scriptsize ±0.6} & \underline{36.50 \scriptsize ±4.1} & 91.49 \scriptsize ±0.4 & 43.27 \scriptsize ±3.1 & 94.57 \scriptsize ±0.3 & 29.17 \scriptsize ±3.0 \\
 &  & Energy & 82.05 \scriptsize ±0.6 & 69.79 \scriptsize ±3.9 & 92.06 & 35.32 & 81.96 \scriptsize ±0.7 & 68.70 \scriptsize ±4.2 & \underline{92.15 \scriptsize ±0.6} & 38.62 \scriptsize ±4.9 & 90.92 \scriptsize ±0.4 & 46.28 \scriptsize ±3.3 & 94.13 \scriptsize ±0.4 & 31.70 \scriptsize ±2.8 \\
 &  & Gradnorm & 60.17 \scriptsize ±1.5 & 87.88 \scriptsize ±2.5 & 85.22 & 44.41 & 62.90 \scriptsize ±0.5 & 86.89 \scriptsize ±0.8 & 81.11 \scriptsize ±1.7 & 59.23 \scriptsize ±3.3 & 81.09 \scriptsize ±1.8 & 57.80 \scriptsize ±2.7 & 91.00 \scriptsize ±1.8 & 34.46 \scriptsize ±3.9 \\
 &  & ViM & 80.62 \scriptsize ±0.7 & 78.13 \scriptsize ±2.3 & 92.34 & 38.14 & 78.90 \scriptsize ±0.8 & 80.30 \scriptsize ±2.2 & 90.54 \scriptsize ±0.7 & 54.70 \scriptsize ±5.0 & 91.87 \scriptsize ±1.2 & 43.84 \scriptsize ±5.6 & 90.13 \scriptsize ±1.8 & 56.97 \scriptsize ±8.5 \\
 &  & Mahal & 49.96 \scriptsize ±2.0 & 96.36 \scriptsize ±0.9 & 61.66 & 78.92 & 46.57 \scriptsize ±1.5 & 96.83 \scriptsize ±0.5 & 50.34 \scriptsize ±1.6 & 95.01 \scriptsize ±0.6 & 63.66 \scriptsize ±3.7 & 86.30 \scriptsize ±2.0 & 47.42 \scriptsize ±6.5 & 96.99 \scriptsize ±0.8 \\
\bottomrule
\end{tabular}
}
\resizebox{.85\textwidth}{!}{
\begin{tabular}{lllbbllllllllll}
\toprule
 &  &  & \multicolumn{2}{c}{\textbf{ID\xmark}} & \multicolumn{2}{c}{\textbf{Textures}} & \multicolumn{2}{c}{\textbf{Colonoscopy}} & \multicolumn{2}{c}{\textbf{Colorectal}} & \multicolumn{2}{c}{\textbf{Noise}} & \multicolumn{2}{c}{\textbf{ImageNet-O}} \\
\textbf{Model} &  & \textbf{Method} & \%AUROC$\uparrow$ & \%FPR@95$\downarrow$ & \%AUROC$\uparrow$ & \%FPR@95$\downarrow$ & \%AUROC$\uparrow$ & \%FPR@95$\downarrow$ & \%AUROC$\uparrow$ & \%FPR@95$\downarrow$ & \%AUROC$\uparrow$ & \%FPR@95$\downarrow$ & \%AUROC$\uparrow$ & \%FPR@95$\downarrow$ \\

\cmidrule(r){1-5} \cmidrule{6-15}

\multirow[c]{16}{*}{\begin{sideways}\shortstack[l]{\textbf{ResNet-50} \\ ID \%Error: 19.01}\end{sideways}} & \multirow[c]{6}{*}{\begin{sideways}\textbf{SIRC}\end{sideways}} & (MSP,$||\b z||_1$) & 90.34 \scriptsize ±0.2 & 52.70 \scriptsize ±3.2 & 93.64 \scriptsize ±0.7 & 32.02 \scriptsize ±3.3 & 95.93 \scriptsize ±1.0 & 25.33 \scriptsize ±6.4 & 95.84 \scriptsize ±3.3 & 24.39 \scriptsize ±13.7 & 90.72 \scriptsize ±6.0 & 49.63 \scriptsize ±20.8 & 83.44 \scriptsize ±0.9 & 58.91 \scriptsize ±1.9 \\
 &  & (MSP,Res.) & \textbf{90.43 \scriptsize ±0.3} & \underline{52.10 \scriptsize ±3.0} & 96.00 \scriptsize ±0.5 & 19.81 \scriptsize ±2.1 & 95.52 \scriptsize ±0.7 & 27.31 \scriptsize ±5.3 & 95.32 \scriptsize ±4.0 & 26.97 \scriptsize ±17.5 & 98.21 \scriptsize ±2.3 & 10.97 \scriptsize ±16.7 & 84.62 \scriptsize ±0.9 & 53.99 \scriptsize ±1.4 \\
 &  & (DR,$||\b z||_1$) & 90.29 \scriptsize ±0.3 & 52.54 \scriptsize ±3.4 & 94.01 \scriptsize ±0.7 & 28.62 \scriptsize ±2.6 & 96.34 \scriptsize ±1.0 & 20.94 \scriptsize ±6.0 & 96.28 \scriptsize ±3.2 & 20.30 \scriptsize ±13.5 & 91.08 \scriptsize ±5.8 & 47.75 \scriptsize ±16.6 & 83.64 \scriptsize ±1.0 & 56.68 \scriptsize ±1.5 \\
 &  & (DR,Res.) & \underline{90.40 \scriptsize ±0.4} & \textbf{51.81 \scriptsize ±2.9} & \underline{96.28 \scriptsize ±0.5} & \underline{17.29 \scriptsize ±2.0} & 95.82 \scriptsize ±0.6 & 23.07 \scriptsize ±4.7 & 95.62 \scriptsize ±4.1 & 23.40 \scriptsize ±18.5 & \underline{98.63 \scriptsize ±1.9} & \underline{7.23 \scriptsize ±10.4} & \underline{84.90 \scriptsize ±0.9} & \underline{51.05 \scriptsize ±1.7} \\
 &  & ($-\mathcal{H}$,$||\b z||_1$) & 90.00 \scriptsize ±0.4 & 54.26 \scriptsize ±2.7 & 94.38 \scriptsize ±0.7 & 27.38 \scriptsize ±2.7 & \underline{96.97 \scriptsize ±0.8} & \underline{16.87 \scriptsize ±4.4} & 96.71 \scriptsize ±2.8 & 18.71 \scriptsize ±13.5 & 91.74 \scriptsize ±4.3 & 45.84 \scriptsize ±17.6 & 84.01 \scriptsize ±0.9 & 56.34 \scriptsize ±2.4 \\
 &  & ($-\mathcal{H}$,Res.) & 90.13 \scriptsize ±0.4 & 54.01 \scriptsize ±3.2 & \underline{96.68 \scriptsize ±0.5} & \underline{15.70 \scriptsize ±2.1} & 96.72 \scriptsize ±0.6 & 18.10 \scriptsize ±3.7 & 96.41 \scriptsize ±3.6 & 20.42 \scriptsize ±16.7 & \underline{99.02 \scriptsize ±1.5} & \underline{4.89 \scriptsize ±5.5} & \underline{85.33 \scriptsize ±0.9} & \underline{50.81 \scriptsize ±2.9} \\

 \cmidrule(r){2-5} \cmidrule{6-15}

& \multirow[c]{10}{*}{} & MSP & \underline{90.41 \scriptsize ±0.3} & 52.13 \scriptsize ±2.0 & 92.88 \scriptsize ±0.8 & 36.61 \scriptsize ±3.1 & 95.75 \scriptsize ±0.8 & 26.52 \scriptsize ±6.2 & 94.86 \scriptsize ±3.5 & 30.28 \scriptsize ±13.6 & 89.33 \scriptsize ±5.7 & 56.83 \scriptsize ±20.2 & 83.29 \scriptsize ±0.9 & 59.78 \scriptsize ±2.1 \\
 &  & DOCTOR & 90.39 \scriptsize ±0.3 & \underline{51.87 \scriptsize ±1.6} & 93.16 \scriptsize ±0.8 & 33.46 \scriptsize ±3.6 & 96.14 \scriptsize ±0.8 & 22.07 \scriptsize ±5.3 & 95.16 \scriptsize ±3.5 & 27.21 \scriptsize ±14.2 & 89.51 \scriptsize ±5.5 & 54.83 \scriptsize ±20.4 & 83.47 \scriptsize ±0.9 & 57.64 \scriptsize ±1.9 \\
 &  & $-\mathcal{H}$ & 90.07 \scriptsize ±0.4 & 54.05 \scriptsize ±2.9 & 93.77 \scriptsize ±0.8 & 30.79 \scriptsize ±3.7 & 96.87 \scriptsize ±0.7 & 17.55 \scriptsize ±4.6 & 95.93 \scriptsize ±3.2 & 23.43 \scriptsize ±14.6 & 90.47 \scriptsize ±4.2 & 51.63 \scriptsize ±19.7 & 83.89 \scriptsize ±0.9 & 57.02 \scriptsize ±1.7 \\
 &  & $||\b z||_1$ & 48.06 \scriptsize ±1.1 & 94.70 \scriptsize ±1.4 & 88.90 \scriptsize ±1.5 & 39.67 \scriptsize ±2.6 & 76.97 \scriptsize ±9.7 & 82.24 \scriptsize ±14.3 & 97.28 \scriptsize ±2.3 & 14.64 \scriptsize ±13.9 & 97.36 \scriptsize ±4.6 & 13.51 \scriptsize ±33.1 & 63.00 \scriptsize ±1.7 & 84.82 \scriptsize ±1.6 \\
 &  & Residual & 47.59 \scriptsize ±1.8 & 96.45 \scriptsize ±1.1 & 82.84 \scriptsize ±2.4 & 46.63 \scriptsize ±3.8 & 38.09 \scriptsize ±13.9 & 99.64 \scriptsize ±0.4 & 53.93 \scriptsize ±13.2 & 88.78 \scriptsize ±10.1 & 91.31 \scriptsize ±6.4 & 20.92 \scriptsize ±12.5 & 68.04 \scriptsize ±2.7 & 78.98 \scriptsize ±2.2 \\
 &  & Max Logit & 83.21 \scriptsize ±0.6 & 65.16 \scriptsize ±3.4 & 95.44 \scriptsize ±0.8 & 22.04 \scriptsize ±2.8 & \textbf{97.65 \scriptsize ±0.7} & \textbf{13.56 \scriptsize ±6.0} & \underline{98.93 \scriptsize ±1.0} & \underline{5.83 \scriptsize ±6.4} & 94.73 \scriptsize ±5.3 & 31.53 \scriptsize ±28.8 & 82.98 \scriptsize ±0.9 & 60.09 \scriptsize ±2.6 \\
 &  & Energy & 82.05 \scriptsize ±0.6 & 69.79 \scriptsize ±3.9 & 95.37 \scriptsize ±0.8 & 22.50 \scriptsize ±2.8 & \underline{97.51 \scriptsize ±0.8} & \underline{14.19 \scriptsize ±6.5} & \textbf{99.07 \scriptsize ±1.0} & \underline{5.00 \scriptsize ±6.5} & 94.93 \scriptsize ±5.4 & 29.05 \scriptsize ±30.8 & 82.52 \scriptsize ±0.9 & 61.86 \scriptsize ±2.7 \\
 &  & Gradnorm & 60.17 \scriptsize ±1.5 & 87.88 \scriptsize ±2.5 & 93.00 \scriptsize ±1.1 & 26.57 \scriptsize ±3.0 & 90.54 \scriptsize ±4.6 & 42.85 \scriptsize ±16.2 & \underline{98.98 \scriptsize ±1.1} & \textbf{4.98 \scriptsize ±7.2} & 97.59 \scriptsize ±4.2 & 13.05 \scriptsize ±31.9 & 70.78 \scriptsize ±1.8 & 73.88 \scriptsize ±3.1 \\
 &  & ViM & 80.62 \scriptsize ±0.7 & 78.13 \scriptsize ±2.3 & \textbf{98.46 \scriptsize ±0.4} & \textbf{7.62 \scriptsize ±2.1} & 94.42 \scriptsize ±1.4 & 44.55 \scriptsize ±14.5 & 98.04 \scriptsize ±1.3 & 8.84 \scriptsize ±10.2 & \textbf{99.82 \scriptsize ±0.1} & \textbf{0.31 \scriptsize ±0.3} & \textbf{88.85 \scriptsize ±0.9} & \textbf{46.15 \scriptsize ±3.9} \\
 &  & Mahal & 49.96 \scriptsize ±2.0 & 96.36 \scriptsize ±0.9 & 84.64 \scriptsize ±2.1 & 46.98 \scriptsize ±3.2 & 41.02 \scriptsize ±14.2 & 99.70 \scriptsize ±0.3 & 57.88 \scriptsize ±12.5 & 88.37 \scriptsize ±8.1 & 94.08 \scriptsize ±5.4 & 20.45 \scriptsize ±13.0 & 69.29 \scriptsize ±2.5 & 79.65 \scriptsize ±1.9 \\
\bottomrule
\end{tabular}
}

\resizebox{.85\textwidth}{!}{
\begin{tabular}{lllbbaallllllll}

\toprule
 &  &  & \multicolumn{2}{c}{\textbf{ID\xmark}} & \multicolumn{2}{c}{\textbf{OOD mean}} & \multicolumn{2}{c}{\textbf{Near-IN-200}} & \multicolumn{2}{c}{\textbf{Caltech-45}} & \multicolumn{2}{c}{\textbf{Openimage-O}} & \multicolumn{2}{c}{\textbf{iNaturalist}} \\
\textbf{Model} &  & \textbf{Method}&\%AUROC$\uparrow$ & \%FPR@95$\downarrow$ & \%AUROC$\uparrow$ & \%FPR@95$\downarrow$ & \%AUROC$\uparrow$ & \%FPR@95$\downarrow$ & \%AUROC$\uparrow$ & \%FPR@95$\downarrow$ & \%AUROC$\uparrow$ & \%FPR@95$\downarrow$ & \%AUROC$\uparrow$ & \%FPR@95$\downarrow$ \\
\cmidrule(r){1-5} \cmidrule{6-15}
\multirow[c]{16}{*}{\begin{sideways}\shortstack[l]{\textbf{MobileNetV2} \\ ID \%Error: 21.35}\end{sideways}} & \multirow[c]{6}{*}{\begin{sideways}\textbf{SIRC}\end{sideways}} & (MSP,$||\b z||_1$) & 89.53 \scriptsize ±0.3 & 55.51 \scriptsize ±1.0 & 92.27 & 34.82 & 84.78 \scriptsize ±0.3 & 61.33 \scriptsize ±1.1 & 90.46 \scriptsize ±0.3 & 43.03 \scriptsize ±0.9 & 91.27 \scriptsize ±0.4 & 44.05 \scriptsize ±1.5 & 94.20 \scriptsize ±0.8 & \underline{31.88 \scriptsize ±3.7} \\
 &  & (MSP,Res.) & \textbf{89.67 \scriptsize ±0.3} & \underline{55.10 \scriptsize ±1.4} & 91.78 & 38.56 & 84.84 \scriptsize ±0.4 & 61.18 \scriptsize ±0.3 & 90.25 \scriptsize ±0.4 & 44.42 \scriptsize ±2.3 & 91.20 \scriptsize ±0.5 & 44.83 \scriptsize ±1.8 & 93.22 \scriptsize ±0.9 & 37.97 \scriptsize ±3.8 \\
 &  & (DR,$||\b z||_1$) & 89.40 \scriptsize ±0.2 & 56.49 \scriptsize ±2.2 & \underline{92.66} & \underline{32.30} & 84.90 \scriptsize ±0.3 & 61.26 \scriptsize ±0.8 & 90.82 \scriptsize ±0.3 & \underline{40.52 \scriptsize ±2.2} & \underline{91.61 \scriptsize ±0.4} & \textbf{42.36 \scriptsize ±1.0} & \underline{94.63 \scriptsize ±0.7} & \textbf{29.15 \scriptsize ±2.8} \\
 &  & (DR,Res.) & \underline{89.60 \scriptsize ±0.3} & 55.69 \scriptsize ±2.0 & 92.08 & 36.21 & 84.98 \scriptsize ±0.3 & \underline{60.92 \scriptsize ±1.1} & 90.58 \scriptsize ±0.4 & 41.98 \scriptsize ±2.3 & 91.51 \scriptsize ±0.5 & \underline{43.22 \scriptsize ±1.9} & 93.40 \scriptsize ±0.9 & 36.31 \scriptsize ±4.2 \\
 &  & ($-\mathcal{H}$,$||\b z||_1$) & 88.90 \scriptsize ±0.2 & 58.64 \scriptsize ±2.1 & \textbf{92.92} & \textbf{32.16} & 84.96 \scriptsize ±0.2 & 62.72 \scriptsize ±0.9 & \underline{91.35 \scriptsize ±0.3} & \textbf{39.89 \scriptsize ±2.1} & \underline{91.82 \scriptsize ±0.4} & 43.99 \scriptsize ±1.4 & \textbf{94.74 \scriptsize ±0.7} & \underline{30.47 \scriptsize ±3.9} \\
 &  & ($-\mathcal{H}$,Res.) & 89.12 \scriptsize ±0.3 & 57.85 \scriptsize ±3.1 & \underline{92.69} & \underline{34.20} & \textbf{85.08 \scriptsize ±0.2} & 62.06 \scriptsize ±0.8 & \underline{91.33 \scriptsize ±0.3} & \underline{39.93 \scriptsize ±2.2} & \textbf{91.93 \scriptsize ±0.4} & \underline{43.80 \scriptsize ±1.2} & 94.01 \scriptsize ±0.8 & 35.24 \scriptsize ±3.7 \\

 \cmidrule(r){2-5} \cmidrule{6-15}

 & \multirow[c]{10}{*}{} & MSP & \underline{89.64 \scriptsize ±0.3} & \textbf{55.03 \scriptsize ±1.5} & 91.54 & 39.73 & 84.84 \scriptsize ±0.3 & \underline{61.03 \scriptsize ±0.9} & 90.17 \scriptsize ±0.3 & 44.77 \scriptsize ±1.7 & 90.91 \scriptsize ±0.5 & 46.34 \scriptsize ±1.8 & 93.57 \scriptsize ±0.9 & 35.85 \scriptsize ±3.8 \\
 &  & DOCTOR & 89.57 \scriptsize ±0.2 & \underline{55.48 \scriptsize ±2.1} & 91.86 & 37.43 & \underline{84.99 \scriptsize ±0.3} & \textbf{60.61 \scriptsize ±0.3} & 90.52 \scriptsize ±0.3 & 42.46 \scriptsize ±2.1 & 91.20 \scriptsize ±0.5 & 44.96 \scriptsize ±1.0 & 93.91 \scriptsize ±0.8 & 33.40 \scriptsize ±3.4 \\
 &  & $-\mathcal{H}$ & 89.02 \scriptsize ±0.2 & 58.43 \scriptsize ±2.2 & 92.37 & 36.04 & \underline{85.03 \scriptsize ±0.2} & 62.53 \scriptsize ±0.4 & 91.16 \scriptsize ±0.3 & 41.27 \scriptsize ±2.0 & 91.54 \scriptsize ±0.4 & 46.11 \scriptsize ±1.4 & \underline{94.24 \scriptsize ±0.7} & 33.85 \scriptsize ±3.8 \\
 &  & $||\b z||_1$ & 53.56 \scriptsize ±0.7 & 93.40 \scriptsize ±0.4 & 81.06 & 53.50 & 56.05 \scriptsize ±0.7 & 92.65 \scriptsize ±0.5 & 75.15 \scriptsize ±1.4 & 73.17 \scriptsize ±2.2 & 74.05 \scriptsize ±1.4 & 68.93 \scriptsize ±1.7 & 86.03 \scriptsize ±1.7 & 48.35 \scriptsize ±5.0 \\
 &  & Residual & 41.99 \scriptsize ±0.8 & 97.30 \scriptsize ±0.3 & 41.42 & 94.11 & 42.46 \scriptsize ±0.7 & 97.37 \scriptsize ±0.3 & 40.90 \scriptsize ±1.2 & 96.70 \scriptsize ±0.9 & 44.63 \scriptsize ±1.1 & 94.39 \scriptsize ±0.6 & 22.88 \scriptsize ±4.7 & 99.18 \scriptsize ±0.4 \\
 &  & Max Logit & 83.14 \scriptsize ±0.6 & 63.85 \scriptsize ±1.8 & 92.08 & 34.64 & 81.75 \scriptsize ±0.4 & 67.36 \scriptsize ±1.3 & \textbf{91.40 \scriptsize ±0.2} & 42.44 \scriptsize ±2.1 & 89.70 \scriptsize ±0.8 & 50.66 \scriptsize ±3.0 & 92.63 \scriptsize ±1.0 & 39.76 \scriptsize ±4.9 \\
 &  & Energy & 81.87 \scriptsize ±0.7 & 67.98 \scriptsize ±2.0 & 91.68 & 36.68 & 80.87 \scriptsize ±0.4 & 70.81 \scriptsize ±1.2 & 90.93 \scriptsize ±0.3 & 45.77 \scriptsize ±1.9 & 88.86 \scriptsize ±0.8 & 54.53 \scriptsize ±3.1 & 91.76 \scriptsize ±1.0 & 44.23 \scriptsize ±5.7 \\
 &  & Gradnorm & 65.27 \scriptsize ±1.1 & 85.73 \scriptsize ±1.1 & 87.25 & 40.67 & 66.07 \scriptsize ±0.7 & 85.13 \scriptsize ±1.0 & 83.94 \scriptsize ±1.0 & 56.57 \scriptsize ±2.9 & 81.94 \scriptsize ±1.2 & 58.20 \scriptsize ±1.7 & 90.73 \scriptsize ±1.3 & 36.80 \scriptsize ±3.4 \\
 &  & ViM & 80.21 \scriptsize ±0.4 & 74.36 \scriptsize ±2.1 & 89.46 & 51.97 & 79.15 \scriptsize ±0.3 & 75.78 \scriptsize ±1.4 & 89.17 \scriptsize ±0.4 & 58.45 \scriptsize ±0.4 & 87.66 \scriptsize ±1.0 & 59.54 \scriptsize ±2.1 & 81.93 \scriptsize ±3.0 & 81.71 \scriptsize ±6.4 \\
 &  & Mahal & 44.44 \scriptsize ±1.0 & 97.14 \scriptsize ±0.6 & 43.65 & 94.20 & 44.57 \scriptsize ±0.7 & 97.23 \scriptsize ±0.4 & 42.82 \scriptsize ±1.1 & 96.64 \scriptsize ±0.8 & 48.03 \scriptsize ±1.2 & 94.11 \scriptsize ±0.8 & 27.31 \scriptsize ±4.8 & 99.07 \scriptsize ±0.3 \\
\bottomrule
\end{tabular}
}
\resizebox{.85\textwidth}{!}{\begin{tabular}{lllbbllllllllll}
\toprule
 &  &  & \multicolumn{2}{c}{\textbf{ID\xmark}} & \multicolumn{2}{c}{\textbf{Textures}} & \multicolumn{2}{c}{\textbf{Colonoscopy}} & \multicolumn{2}{c}{\textbf{Colorectal}} & \multicolumn{2}{c}{\textbf{Noise}} & \multicolumn{2}{c}{\textbf{ImageNet-O}} \\
\textbf{Model} &  & \textbf{Method} & \%AUROC$\uparrow$ & \%FPR@95$\downarrow$ & \%AUROC$\uparrow$ & \%FPR@95$\downarrow$ & \%AUROC$\uparrow$ & \%FPR@95$\downarrow$ & \%AUROC$\uparrow$ & \%FPR@95$\downarrow$ & \%AUROC$\uparrow$ & \%FPR@95$\downarrow$ & \%AUROC$\uparrow$ & \%FPR@95$\downarrow$ \\

\cmidrule(r){1-5} \cmidrule{6-15}

\multirow[c]{16}{*}{\begin{sideways}\shortstack[l]{\textbf{MobileNetV2} \\ ID \%Error: 21.35}\end{sideways}} & \multirow[c]{6}{*}{\begin{sideways}\textbf{SIRC}\end{sideways}} & (MSP,$||\b z||_1$) & 89.53 \scriptsize ±0.3 & 55.51 \scriptsize ±1.0 & 94.05 \scriptsize ±0.4 & 28.69 \scriptsize ±0.5 & 96.64 \scriptsize ±0.8 & 19.38 \scriptsize ±4.8 & 96.98 \scriptsize ±1.4 & 19.39 \scriptsize ±8.8 & 98.77 \scriptsize ±1.1 & 7.61 \scriptsize ±7.7 & 83.30 \scriptsize ±0.3 & 57.98 \scriptsize ±1.2 \\
 &  & (MSP,Res.) & \textbf{89.67 \scriptsize ±0.3} & \underline{55.10 \scriptsize ±1.4} & 94.26 \scriptsize ±0.2 & 28.37 \scriptsize ±1.2 & 96.25 \scriptsize ±0.9 & 21.57 \scriptsize ±4.3 & 95.45 \scriptsize ±1.7 & 29.21 \scriptsize ±9.2 & 96.45 \scriptsize ±2.5 & 24.49 \scriptsize ±24.8 & 84.09 \scriptsize ±0.4 & 55.02 \scriptsize ±1.7 \\
 &  & (DR,$||\b z||_1$) & 89.40 \scriptsize ±0.2 & 56.49 \scriptsize ±2.2 & 94.54 \scriptsize ±0.4 & 25.70 \scriptsize ±1.1 & 97.07 \scriptsize ±0.7 & 15.64 \scriptsize ±3.1 & 97.61 \scriptsize ±1.3 & 14.88 \scriptsize ±7.2 & \underline{99.25 \scriptsize ±0.9} & 4.38 \scriptsize ±5.6 & 83.56 \scriptsize ±0.2 & 56.83 \scriptsize ±1.6 \\
 &  & (DR,Res.) & \underline{89.60 \scriptsize ±0.3} & 55.69 \scriptsize ±2.0 & 94.68 \scriptsize ±0.2 & 25.37 \scriptsize ±0.7 & 96.61 \scriptsize ±0.8 & 17.89 \scriptsize ±3.8 & 95.77 \scriptsize ±1.6 & 26.30 \scriptsize ±9.1 & 96.68 \scriptsize ±2.7 & 21.41 \scriptsize ±30.2 & \underline{84.49 \scriptsize ±0.4} & \underline{52.49 \scriptsize ±2.2} \\
 &  & ($-\mathcal{H}$,$||\b z||_1$) & 88.90 \scriptsize ±0.2 & 58.64 \scriptsize ±2.1 & 94.88 \scriptsize ±0.4 & \underline{25.26 \scriptsize ±1.1} & \underline{97.71 \scriptsize ±0.5} & \underline{11.67 \scriptsize ±3.3} & 97.82 \scriptsize ±1.1 & 13.78 \scriptsize ±7.6 & 99.07 \scriptsize ±1.2 & 4.28 \scriptsize ±6.6 & 83.93 \scriptsize ±0.3 & 57.35 \scriptsize ±1.6 \\
 &  & ($-\mathcal{H}$,Res.) & 89.12 \scriptsize ±0.3 & 57.85 \scriptsize ±3.1 & \underline{95.37 \scriptsize ±0.2} & \underline{22.62 \scriptsize ±0.8} & 97.57 \scriptsize ±0.6 & 12.44 \scriptsize ±3.8 & 96.76 \scriptsize ±1.4 & 21.12 \scriptsize ±10.8 & 97.29 \scriptsize ±2.8 & 17.55 \scriptsize ±30.2 & \underline{84.91 \scriptsize ±0.3} & \underline{53.07 \scriptsize ±1.2} \\

 \cmidrule(r){2-5} \cmidrule{6-15}
 & \multirow[c]{10}{*}{} & MSP & \underline{89.64 \scriptsize ±0.3} & \textbf{55.03 \scriptsize ±1.5} & 92.93 \scriptsize ±0.5 & 35.28 \scriptsize ±0.8 & 96.51 \scriptsize ±0.8 & 20.36 \scriptsize ±4.6 & 95.63 \scriptsize ±1.6 & 28.52 \scriptsize ±9.4 & 96.22 \scriptsize ±2.7 & 26.33 \scriptsize ±25.6 & 83.11 \scriptsize ±0.3 & 59.05 \scriptsize ±1.3 \\
 &  & DOCTOR & 89.57 \scriptsize ±0.2 & \underline{55.48 \scriptsize ±2.1} & 93.32 \scriptsize ±0.4 & 32.59 \scriptsize ±1.6 & 96.99 \scriptsize ±0.8 & 16.46 \scriptsize ±3.8 & 96.04 \scriptsize ±1.5 & 25.23 \scriptsize ±8.9 & 96.46 \scriptsize ±2.9 & 23.31 \scriptsize ±31.4 & 83.34 \scriptsize ±0.3 & 57.87 \scriptsize ±1.3 \\
 &  & $-\mathcal{H}$ & 89.02 \scriptsize ±0.2 & 58.43 \scriptsize ±2.2 & 94.05 \scriptsize ±0.4 & 29.72 \scriptsize ±0.9 & 97.68 \scriptsize ±0.6 & 12.13 \scriptsize ±3.5 & 96.82 \scriptsize ±1.4 & 21.15 \scriptsize ±10.2 & 97.07 \scriptsize ±2.8 & 19.33 \scriptsize ±30.7 & 83.77 \scriptsize ±0.3 & 58.30 \scriptsize ±1.2 \\
 &  & $||\b z||_1$ & 53.56 \scriptsize ±0.7 & 93.40 \scriptsize ±0.4 & 92.88 \scriptsize ±0.3 & 27.55 \scriptsize ±1.5 & 79.90 \scriptsize ±5.7 & 80.70 \scriptsize ±10.1 & 98.20 \scriptsize ±1.1 & \underline{9.40 \scriptsize ±6.4} & \underline{99.93 \scriptsize ±0.0} & \underline{0.01 \scriptsize ±0.0} & 67.33 \scriptsize ±2.0 & 80.74 \scriptsize ±3.0 \\
 &  & Residual & 41.99 \scriptsize ±0.8 & 97.30 \scriptsize ±0.3 & 56.86 \scriptsize ±1.4 & 78.82 \scriptsize ±2.1 & 27.32 \scriptsize ±6.6 & 99.68 \scriptsize ±1.1 & 28.41 \scriptsize ±8.1 & 99.38 \scriptsize ±1.2 & 49.61 \scriptsize ±18.3 & 93.66 \scriptsize ±5.7 & 59.77 \scriptsize ±1.2 & 87.83 \scriptsize ±0.7 \\
 &  & Max Logit & 83.14 \scriptsize ±0.6 & 63.85 \scriptsize ±1.8 & 95.13 \scriptsize ±0.4 & 25.45 \scriptsize ±1.1 & \textbf{98.12 \scriptsize ±0.5} & \textbf{9.64 \scriptsize ±3.7} & \underline{98.27 \scriptsize ±1.0} & 9.72 \scriptsize ±7.5 & 98.74 \scriptsize ±0.9 & 3.42 \scriptsize ±5.1 & 83.02 \scriptsize ±0.6 & 63.26 \scriptsize ±1.5 \\
 &  & Energy & 81.87 \scriptsize ±0.7 & 67.98 \scriptsize ±2.0 & 95.02 \scriptsize ±0.4 & 26.47 \scriptsize ±2.4 & \underline{97.99 \scriptsize ±0.5} & \underline{10.64 \scriptsize ±4.2} & \underline{98.43 \scriptsize ±1.1} & \underline{8.65 \scriptsize ±8.3} & 98.93 \scriptsize ±0.8 & \underline{2.55 \scriptsize ±4.4} & 82.36 \scriptsize ±0.7 & 66.51 \scriptsize ±1.2 \\
 &  & Gradnorm & 65.27 \scriptsize ±1.1 & 85.73 \scriptsize ±1.1 & \textbf{95.47 \scriptsize ±0.2} & \textbf{19.24 \scriptsize ±1.3} & 93.35 \scriptsize ±2.2 & 36.23 \scriptsize ±10.0 & \textbf{99.25 \scriptsize ±0.6} & \textbf{2.98 \scriptsize ±3.2} & \textbf{99.95 \scriptsize ±0.0} & \textbf{0.00 \scriptsize ±0.0} & 74.50 \scriptsize ±1.6 & 70.89 \scriptsize ±2.3 \\
 &  & ViM & 80.21 \scriptsize ±0.4 & 74.36 \scriptsize ±2.1 & \underline{95.46 \scriptsize ±0.2} & 26.00 \scriptsize ±1.6 & 93.62 \scriptsize ±1.5 & 50.41 \scriptsize ±13.5 & 94.04 \scriptsize ±2.1 & 43.82 \scriptsize ±16.7 & 96.71 \scriptsize ±0.5 & 19.97 \scriptsize ±9.9 & \textbf{87.44 \scriptsize ±0.6} & \textbf{52.09 \scriptsize ±2.0} \\
 &  & Mahal & 44.44 \scriptsize ±1.0 & 97.14 \scriptsize ±0.6 & 58.12 \scriptsize ±1.6 & 79.53 \scriptsize ±2.3 & 29.12 \scriptsize ±7.4 & 99.67 \scriptsize ±1.0 & 30.35 \scriptsize ±8.1 & 99.15 \scriptsize ±2.2 & 51.22 \scriptsize ±15.5 & 94.62 \scriptsize ±5.4 & 61.28 \scriptsize ±1.1 & 87.74 \scriptsize ±1.1 \\

\bottomrule
\end{tabular}
}

\resizebox{.85\textwidth}{!}{
\begin{tabular}{lllbbaallllllll}

\toprule
 &  &  & \multicolumn{2}{c}{\textbf{ID\xmark}} & \multicolumn{2}{c}{\textbf{OOD mean}} & \multicolumn{2}{c}{\textbf{Near-IN-200}} & \multicolumn{2}{c}{\textbf{Caltech-45}} & \multicolumn{2}{c}{\textbf{Openimage-O}} & \multicolumn{2}{c}{\textbf{iNaturalist}} \\
\textbf{Model} &  & \textbf{Method}&\%AUROC$\uparrow$ & \%FPR@95$\downarrow$ & \%AUROC$\uparrow$ & \%FPR@95$\downarrow$ & \%AUROC$\uparrow$ & \%FPR@95$\downarrow$ & \%AUROC$\uparrow$ & \%FPR@95$\downarrow$ & \%AUROC$\uparrow$ & \%FPR@95$\downarrow$ & \%AUROC$\uparrow$ & \%FPR@95$\downarrow$ \\
\cmidrule(r){1-5} \cmidrule{6-15}

\multirow[c]{16}{*}{\begin{sideways}\shortstack[l]{\textbf{DenseNet-121} \\ ID \%Error: 17.20}\end{sideways}} & \multirow[c]{6}{*}{\begin{sideways}\textbf{SIRC}\end{sideways}} & (MSP,$||\b z||_1$) & \underline{90.22 \scriptsize ±0.8} & 52.41 \scriptsize ±2.7 & 91.68 & 38.83 & 85.28 \scriptsize ±0.1 & 59.35 \scriptsize ±1.0 & 91.33 \scriptsize ±0.5 & 40.80 \scriptsize ±2.0 & 91.88 \scriptsize ±0.4 & 43.34 \scriptsize ±1.4 & 93.52 \scriptsize ±0.8 & 35.88 \scriptsize ±2.8 \\
 &  & (MSP,Res.) & 90.20 \scriptsize ±0.8 & 52.42 \scriptsize ±4.2 & \underline{92.81} & \underline{32.68} & 85.18 \scriptsize ±0.1 & 59.65 \scriptsize ±1.7 & 91.29 \scriptsize ±0.5 & 40.51 \scriptsize ±1.6 & \underline{92.55 \scriptsize ±0.3} & 39.24 \scriptsize ±1.0 & 93.42 \scriptsize ±0.7 & 35.98 \scriptsize ±1.9 \\
 &  & (DR,$||\b z||_1$) & 90.21 \scriptsize ±0.8 & \underline{52.36 \scriptsize ±2.9} & 91.99 & 35.30 & 85.42 \scriptsize ±0.1 & 57.49 \scriptsize ±1.9 & 91.63 \scriptsize ±0.5 & 37.15 \scriptsize ±1.1 & 92.14 \scriptsize ±0.4 & 39.89 \scriptsize ±1.2 & 93.85 \scriptsize ±0.7 & 31.95 \scriptsize ±2.0 \\
 &  & (DR,Res.) & 90.18 \scriptsize ±0.8 & 52.46 \scriptsize ±4.3 & \underline{93.04} & \underline{29.68} & 85.28 \scriptsize ±0.1 & 57.98 \scriptsize ±2.6 & 91.53 \scriptsize ±0.5 & 36.84 \scriptsize ±1.4 & \underline{92.83 \scriptsize ±0.3} & \underline{35.57 \scriptsize ±1.1} & 93.69 \scriptsize ±0.7 & 32.08 \scriptsize ±1.0 \\
 &  & ($-\mathcal{H}$,$||\b z||_1$) & 89.95 \scriptsize ±0.9 & 53.96 \scriptsize ±2.5 & 92.42 & 32.92 & \underline{85.64 \scriptsize ±0.1} & \textbf{56.98 \scriptsize ±1.5} & \textbf{92.18 \scriptsize ±0.4} & \textbf{34.18 \scriptsize ±1.7} & 92.50 \scriptsize ±0.5 & \underline{38.43 \scriptsize ±2.0} & \textbf{94.16 \scriptsize ±0.7} & \textbf{30.23 \scriptsize ±2.7} \\
 &  & ($-\mathcal{H}$,Res.) & 89.92 \scriptsize ±0.8 & 54.17 \scriptsize ±3.2 & \textbf{93.45} & \textbf{27.97} & \underline{85.50 \scriptsize ±0.1} & 57.90 \scriptsize ±1.8 & \underline{92.13 \scriptsize ±0.4} & \underline{34.47 \scriptsize ±1.1} & \textbf{93.16 \scriptsize ±0.4} & \textbf{34.87 \scriptsize ±1.5} & \underline{94.07 \scriptsize ±0.6} & \underline{30.60 \scriptsize ±1.9} \\
 
 \cmidrule(r){2-5} \cmidrule{6-15}

& \multirow[c]{10}{*}{} & MSP & \underline{90.30 \scriptsize ±0.8} & \underline{51.85 \scriptsize ±2.9} & 91.44 & 40.44 & 85.34 \scriptsize ±0.1 & 59.20 \scriptsize ±1.0 & 91.23 \scriptsize ±0.5 & 41.37 \scriptsize ±2.1 & 91.78 \scriptsize ±0.4 & 43.99 \scriptsize ±1.3 & 93.32 \scriptsize ±0.7 & 37.11 \scriptsize ±2.8 \\
 &  & DOCTOR & \textbf{90.32 \scriptsize ±0.8} & \textbf{51.80 \scriptsize ±3.3} & 91.71 & 36.95 & 85.48 \scriptsize ±0.1 & \underline{57.16 \scriptsize ±1.8} & 91.51 \scriptsize ±0.5 & 37.63 \scriptsize ±1.2 & 92.03 \scriptsize ±0.4 & 40.45 \scriptsize ±1.3 & 93.60 \scriptsize ±0.7 & 33.19 \scriptsize ±1.9 \\
 &  & $-\mathcal{H}$ & 90.04 \scriptsize ±0.9 & 53.41 \scriptsize ±2.3 & 92.24 & 34.49 & \textbf{85.70 \scriptsize ±0.1} & \underline{57.01 \scriptsize ±1.5} & \underline{92.11 \scriptsize ±0.4} & \underline{34.78 \scriptsize ±1.7} & 92.43 \scriptsize ±0.5 & 39.17 \scriptsize ±2.5 & \underline{94.00 \scriptsize ±0.7} & \underline{31.37 \scriptsize ±3.1} \\
 &  & $||\b z||_1$ & 36.87 \scriptsize ±2.3 & 98.70 \scriptsize ±0.4 & 63.53 & 80.35 & 42.10 \scriptsize ±1.9 & 98.50 \scriptsize ±0.6 & 55.63 \scriptsize ±4.5 & 91.89 \scriptsize ±2.1 & 53.48 \scriptsize ±3.7 & 91.83 \scriptsize ±2.4 & 64.95 \scriptsize ±5.9 & 87.01 \scriptsize ±6.2 \\
 &  & Residual & 46.08 \scriptsize ±1.1 & 95.44 \scriptsize ±0.8 & 69.38 & 71.33 & 44.76 \scriptsize ±0.9 & 96.14 \scriptsize ±0.1 & 55.03 \scriptsize ±1.6 & 91.27 \scriptsize ±0.9 & 70.77 \scriptsize ±2.2 & 78.26 \scriptsize ±1.6 & 64.91 \scriptsize ±6.1 & 90.93 \scriptsize ±2.5 \\
 &  & Max Logit & 83.28 \scriptsize ±1.3 & 62.32 \scriptsize ±2.7 & 91.34 & 36.42 & 82.50 \scriptsize ±0.3 & 63.06 \scriptsize ±1.8 & 91.57 \scriptsize ±0.6 & 37.89 \scriptsize ±2.2 & 89.71 \scriptsize ±1.1 & 48.54 \scriptsize ±2.6 & 91.28 \scriptsize ±1.2 & 42.80 \scriptsize ±4.7 \\
 &  & Energy & 82.12 \scriptsize ±1.3 & 66.54 \scriptsize ±3.4 & 90.92 & 38.87 & 81.82 \scriptsize ±0.4 & 66.32 \scriptsize ±1.6 & 91.14 \scriptsize ±0.7 & 40.74 \scriptsize ±2.4 & 88.88 \scriptsize ±1.2 & 52.80 \scriptsize ±3.0 & 90.37 \scriptsize ±1.4 & 47.67 \scriptsize ±5.2 \\
 &  & Gradnorm & 50.18 \scriptsize ±2.6 & 95.19 \scriptsize ±1.6 & 76.18 & 62.58 & 54.43 \scriptsize ±2.0 & 93.93 \scriptsize ±1.9 & 72.05 \scriptsize ±4.0 & 74.98 \scriptsize ±4.7 & 67.42 \scriptsize ±3.6 & 79.78 \scriptsize ±4.3 & 77.17 \scriptsize ±4.8 & 67.53 \scriptsize ±8.2 \\
 &  & ViM & 76.63 \scriptsize ±1.3 & 84.73 \scriptsize ±1.9 & 90.50 & 44.71 & 75.68 \scriptsize ±0.8 & 86.14 \scriptsize ±1.7 & 87.10 \scriptsize ±1.1 & 64.27 \scriptsize ±4.3 & 89.79 \scriptsize ±0.9 & 51.76 \scriptsize ±4.0 & 89.00 \scriptsize ±1.5 & 59.37 \scriptsize ±5.9 \\
 &  & Mahal & 56.97 \scriptsize ±1.1 & 94.57 \scriptsize ±1.1 & 73.14 & 71.91 & 53.97 \scriptsize ±1.1 & 95.39 \scriptsize ±0.3 & 60.81 \scriptsize ±1.5 & 90.95 \scriptsize ±1.0 & 76.58 \scriptsize ±1.6 & 76.51 \scriptsize ±2.6 & 69.67 \scriptsize ±4.7 & 89.79 \scriptsize ±4.0 \\
\bottomrule
\end{tabular}
}
\resizebox{.85\textwidth}{!}{\begin{tabular}{lllbbllllllllll}
\toprule
 &  &  & \multicolumn{2}{c}{\textbf{ID\xmark}} & \multicolumn{2}{c}{\textbf{Textures}} & \multicolumn{2}{c}{\textbf{Colonoscopy}} & \multicolumn{2}{c}{\textbf{Colorectal}} & \multicolumn{2}{c}{\textbf{Noise}} & \multicolumn{2}{c}{\textbf{ImageNet-O}} \\
\textbf{Model} &  & \textbf{Method} & \%AUROC$\uparrow$ & \%FPR@95$\downarrow$ & \%AUROC$\uparrow$ & \%FPR@95$\downarrow$ & \%AUROC$\uparrow$ & \%FPR@95$\downarrow$ & \%AUROC$\uparrow$ & \%FPR@95$\downarrow$ & \%AUROC$\uparrow$ & \%FPR@95$\downarrow$ & \%AUROC$\uparrow$ & \%FPR@95$\downarrow$ \\

\cmidrule(r){1-5} \cmidrule{6-15}

\multirow[c]{16}{*}{\begin{sideways}\shortstack[l]{\textbf{DenseNet-121} \\ ID \%Error: 17.20}\end{sideways}} & \multirow[c]{6}{*}{\begin{sideways}\textbf{SIRC}\end{sideways}} & (MSP,$||\b z||_1$) & \underline{90.22 \scriptsize ±0.8} & 52.41 \scriptsize ±2.7 & 93.68 \scriptsize ±0.2 & 32.48 \scriptsize ±0.9 & 95.38 \scriptsize ±0.9 & 27.10 \scriptsize ±5.5 & 96.44 \scriptsize ±2.3 & 23.54 \scriptsize ±14.1 & 94.31 \scriptsize ±8.3 & 28.79 \scriptsize ±25.0 & 83.32 \scriptsize ±0.7 & 58.19 \scriptsize ±1.4 \\
 &  & (MSP,Res.) & 90.20 \scriptsize ±0.8 & 52.42 \scriptsize ±4.2 & 96.55 \scriptsize ±0.1 & 16.60 \scriptsize ±0.9 & 95.14 \scriptsize ±0.8 & 27.71 \scriptsize ±4.6 & 96.89 \scriptsize ±1.7 & 19.72 \scriptsize ±10.6 & 99.70 \scriptsize ±1.1 & 2.01 \scriptsize ±8.6 & 84.59 \scriptsize ±0.6 & \underline{52.69 \scriptsize ±2.1} \\
 &  & (DR,$||\b z||_1$) & 90.21 \scriptsize ±0.8 & \underline{52.36 \scriptsize ±2.9} & 94.02 \scriptsize ±0.2 & 28.81 \scriptsize ±0.3 & 95.73 \scriptsize ±0.9 & 22.94 \scriptsize ±5.4 & 96.84 \scriptsize ±2.2 & 19.40 \scriptsize ±13.2 & 94.72 \scriptsize ±8.3 & 24.54 \scriptsize ±24.1 & 83.52 \scriptsize ±0.7 & 55.50 \scriptsize ±1.7 \\
 &  & (DR,Res.) & 90.18 \scriptsize ±0.8 & 52.46 \scriptsize ±4.3 & \underline{96.77 \scriptsize ±0.1} & \underline{14.26 \scriptsize ±1.2} & 95.42 \scriptsize ±0.8 & 23.65 \scriptsize ±4.6 & 97.24 \scriptsize ±1.6 & 15.56 \scriptsize ±9.2 & \underline{99.78 \scriptsize ±0.7} & \underline{1.23 \scriptsize ±5.3} & \underline{84.84 \scriptsize ±0.6} & \underline{49.91 \scriptsize ±3.2} \\
 &  & ($-\mathcal{H}$,$||\b z||_1$) & 89.95 \scriptsize ±0.9 & 53.96 \scriptsize ±2.5 & 94.48 \scriptsize ±0.2 & 26.16 \scriptsize ±1.3 & 96.43 \scriptsize ±0.9 & 17.89 \scriptsize ±5.2 & 97.38 \scriptsize ±1.9 & 15.64 \scriptsize ±12.0 & 95.17 \scriptsize ±8.0 & 22.30 \scriptsize ±19.5 & 83.89 \scriptsize ±0.7 & 54.48 \scriptsize ±1.9 \\
 &  & ($-\mathcal{H}$,Res.) & 89.92 \scriptsize ±0.8 & 54.17 \scriptsize ±3.2 & \underline{97.07 \scriptsize ±0.1} & \underline{13.02 \scriptsize ±1.0} & 96.21 \scriptsize ±0.8 & 18.57 \scriptsize ±4.3 & \underline{97.78 \scriptsize ±1.4} & 12.58 \scriptsize ±8.4 & \textbf{99.91 \scriptsize ±0.3} & \underline{0.32 \scriptsize ±1.4} & \underline{85.22 \scriptsize ±0.6} & \textbf{49.44 \scriptsize ±1.4} \\

 \cmidrule(r){2-5} \cmidrule{6-15}
 & \multirow[c]{10}{*}{} & MSP & \underline{90.30 \scriptsize ±0.8} & \underline{51.85 \scriptsize ±2.9} & 93.25 \scriptsize ±0.2 & 35.01 \scriptsize ±0.8 & 95.42 \scriptsize ±0.8 & 26.81 \scriptsize ±5.1 & 95.92 \scriptsize ±2.1 & 26.92 \scriptsize ±13.1 & 93.44 \scriptsize ±8.4 & 35.16 \scriptsize ±28.5 & 83.31 \scriptsize ±0.6 & 58.37 \scriptsize ±1.3 \\
 &  & DOCTOR & \textbf{90.32 \scriptsize ±0.8} & \textbf{51.80 \scriptsize ±3.3} & 93.54 \scriptsize ±0.2 & 31.47 \scriptsize ±0.5 & 95.78 \scriptsize ±0.9 & 22.56 \scriptsize ±4.8 & 96.25 \scriptsize ±2.1 & 23.09 \scriptsize ±12.5 & 93.72 \scriptsize ±8.3 & 31.28 \scriptsize ±29.0 & 83.49 \scriptsize ±0.6 & 55.73 \scriptsize ±1.9 \\
 &  & $-\mathcal{H}$ & 90.04 \scriptsize ±0.9 & 53.41 \scriptsize ±2.3 & 94.14 \scriptsize ±0.2 & 28.14 \scriptsize ±1.7 & \underline{96.47 \scriptsize ±0.8} & \underline{17.71 \scriptsize ±5.2} & 96.99 \scriptsize ±1.9 & 18.30 \scriptsize ±12.3 & 94.43 \scriptsize ±7.8 & 29.18 \scriptsize ±22.9 & 83.89 \scriptsize ±0.7 & 54.76 \scriptsize ±2.3 \\
 &  & $||\b z||_1$ & 36.87 \scriptsize ±2.3 & 98.70 \scriptsize ±0.4 & 75.96 \scriptsize ±4.4 & 69.61 \scriptsize ±9.0 & 47.58 \scriptsize ±11.0 & 97.80 \scriptsize ±0.4 & 87.28 \scriptsize ±9.2 & 60.31 \scriptsize ±32.1 & 95.31 \scriptsize ±5.3 & 29.94 \scriptsize ±34.9 & 49.50 \scriptsize ±4.0 & 96.31 \scriptsize ±1.0 \\
 &  & Residual & 46.08 \scriptsize ±1.1 & 95.44 \scriptsize ±0.8 & 92.00 \scriptsize ±0.4 & 29.64 \scriptsize ±0.9 & 49.87 \scriptsize ±4.5 & 99.46 \scriptsize ±1.0 & 76.95 \scriptsize ±1.7 & 72.88 \scriptsize ±7.1 & 98.07 \scriptsize ±2.1 & 9.12 \scriptsize ±6.3 & 72.05 \scriptsize ±0.9 & 74.29 \scriptsize ±1.4 \\
 &  & Max Logit & 83.28 \scriptsize ±1.3 & 62.32 \scriptsize ±2.7 & 94.08 \scriptsize ±0.7 & 27.50 \scriptsize ±2.1 & \textbf{96.96 \scriptsize ±0.8} & \textbf{15.92 \scriptsize ±4.8} & \underline{98.20 \scriptsize ±1.4} & \underline{10.35 \scriptsize ±10.1} & 96.09 \scriptsize ±6.8 & 20.58 \scriptsize ±18.2 & 81.65 \scriptsize ±1.9 & 61.15 \scriptsize ±1.9 \\
 &  & Energy & 82.12 \scriptsize ±1.3 & 66.54 \scriptsize ±3.4 & 93.82 \scriptsize ±0.8 & 29.16 \scriptsize ±2.8 & \underline{96.80 \scriptsize ±0.9} & \underline{16.88 \scriptsize ±5.1} & \textbf{98.26 \scriptsize ±1.5} & \textbf{9.99 \scriptsize ±10.5} & 96.13 \scriptsize ±6.7 & 21.78 \scriptsize ±20.4 & 81.07 \scriptsize ±2.0 & 64.51 \scriptsize ±2.2 \\
 &  & Gradnorm & 50.18 \scriptsize ±2.6 & 95.19 \scriptsize ±1.6 & 86.17 \scriptsize ±2.9 & 46.07 \scriptsize ±7.8 & 74.52 \scriptsize ±11.0 & 73.99 \scriptsize ±12.1 & 95.79 \scriptsize ±4.2 & 24.50 \scriptsize ±22.9 & 97.25 \scriptsize ±4.5 & 14.87 \scriptsize ±21.6 & 60.78 \scriptsize ±3.7 & 87.58 \scriptsize ±3.6 \\
 &  & ViM & 76.63 \scriptsize ±1.3 & 84.73 \scriptsize ±1.9 & \textbf{98.07 \scriptsize ±0.2} & \textbf{9.82 \scriptsize ±1.4} & 92.00 \scriptsize ±0.8 & 63.03 \scriptsize ±7.1 & 97.70 \scriptsize ±0.5 & \underline{11.13 \scriptsize ±3.7} & \underline{99.85 \scriptsize ±0.1} & \textbf{0.04 \scriptsize ±0.1} & \textbf{85.30 \scriptsize ±1.4} & 56.82 \scriptsize ±3.3 \\
 &  & Mahal & 56.97 \scriptsize ±1.1 & 94.57 \scriptsize ±1.1 & 91.55 \scriptsize ±0.4 & 31.71 \scriptsize ±0.8 & 60.34 \scriptsize ±9.6 & 98.88 \scriptsize ±1.7 & 70.77 \scriptsize ±4.0 & 79.75 \scriptsize ±7.1 & 96.66 \scriptsize ±2.9 & 12.67 \scriptsize ±5.6 & 77.93 \scriptsize ±1.2 & 71.51 \scriptsize ±2.4 \\
\bottomrule

\bottomrule
\end{tabular}
}

\end{table}
\begin{table}[]
    \centering
        \caption{\%AUROC and \%FPR@95 results for single pre-trained ImageNet-1k models.}
    \label{tab:pretrained_imagenet}
        \resizebox{\textwidth}{!}{
    \begin{tabular}{lllbbaallllllllllllll}
\toprule
 &  &  & \multicolumn{2}{c}{\textbf{ID\xmark}} & \multicolumn{2}{c}{\textbf{OOD mean}} & \multicolumn{2}{c}{\textbf{Openimage-O}} & \multicolumn{2}{c}{\textbf{iNaturalist}} & \multicolumn{2}{c}{\textbf{Textures}} & \multicolumn{2}{c}{\textbf{Colonoscopy}} & \multicolumn{2}{c}{\textbf{Colorectal}} & \multicolumn{2}{c}{\textbf{Noise}} & \multicolumn{2}{c}{\textbf{ImageNet-O}} \\
\textbf{Model} &  & \textbf{Method} & AUROC$\uparrow$ & FPR@95$\downarrow$ & AUROC$\uparrow$ & FPR@95$\downarrow$ & AUROC$\uparrow$ & FPR@95$\downarrow$ & AUROC$\uparrow$ & FPR@95$\downarrow$ & AUROC$\uparrow$ & FPR@95$\downarrow$ & AUROC$\uparrow$ & FPR@95$\downarrow$ & AUROC$\uparrow$ & FPR@95$\downarrow$ & AUROC$\uparrow$ & FPR@95$\downarrow$ & AUROC$\uparrow$ & FPR@95$\downarrow$ \\
\cmidrule(r){1-5} \cmidrule{6-21}
\multirow[c]{16}{*}{\begin{sideways}\shortstack[l]{\textbf{ResNetV2-101} \\ ID \%Error: 22.63}\end{sideways}} & \multirow[c]{6}{*}{\begin{sideways}\textbf{SIRC}\end{sideways}} & (MSP,$||\b z||_1$) & \underline{86.17} & \underline{63.37} & 90.08 & 37.09 & 90.25 & 47.09 & 94.37 & 29.13 & 88.74 & 43.84 & 97.10 & 16.69 & 93.94 & 30.28 & 99.26 & 4.32 & 66.93 & 88.30 \\
 &  & (MSP,Res.) & \underline{86.31} & \underline{62.35} & 92.89 & \underline{25.23} & \underline{92.69} & \underline{34.29} & 94.61 & \underline{27.17} & 96.80 & 10.04 & 97.18 & 14.69 & 96.88 & 14.06 & 99.99 & \textbf{0.00} & 72.07 & 76.35 \\
 &  & (DR,$||\b z||_1$) & 85.36 & 66.04 & 90.44 & 35.13 & 90.35 & 47.22 & \underline{94.75} & \underline{26.89} & 89.19 & 41.72 & 97.13 & 15.22 & 94.85 & 23.82 & 99.48 & 2.93 & 67.33 & 88.10 \\
 &  & (DR,Res.) & 85.55 & 64.66 & \underline{93.34} & \textbf{22.76} & \textbf{93.25} & \textbf{31.27} & \textbf{94.99} & \textbf{24.30} & 97.15 & \underline{8.06} & 97.36 & 12.46 & \underline{97.52} & \underline{10.10} & \underline{99.99} & \underline{0.00} & 73.12 & 73.10 \\
 &  & ($-\mathcal{H}$,$||\b z||_1$) & 83.43 & 69.62 & 90.98 & 36.74 & 90.46 & 51.57 & 94.53 & 33.11 & 89.06 & 47.03 & 98.37 & 9.66 & 95.71 & 23.18 & 99.48 & 3.80 & 69.25 & 88.80 \\
 &  & ($-\mathcal{H}$,Res.) & 83.50 & 68.91 & \underline{93.67} & \underline{25.61} & \underline{92.78} & \underline{40.20} & \underline{94.63} & 33.51 & 97.05 & 10.19 & 98.33 & 9.02 & \underline{98.12} & \underline{8.62} & \underline{100.00} & \underline{0.00} & 74.76 & 77.70 \\
  \cmidrule(r){2-5} \cmidrule{6-21}
 & \multirow[c]{10}{*}{} & MSP & \textbf{86.35} & \textbf{61.93} & 89.16 & 41.81 & 90.13 & 47.48 & 93.70 & 32.96 & 87.04 & 51.90 & 97.47 & 14.92 & 91.55 & 44.28 & 97.37 & 12.53 & 66.87 & 88.60 \\
 &  & DOCTOR & 85.67 & 64.49 & 89.57 & 40.48 & 90.33 & 47.64 & 93.95 & 32.04 & 87.27 & 52.64 & 97.89 & 12.61 & 92.34 & 39.78 & 97.94 & 9.74 & 67.25 & 88.90 \\
 &  & $-\mathcal{H}$ & 83.49 & 69.09 & 90.25 & 41.32 & 90.23 & 54.09 & 93.80 & 38.97 & 87.47 & 54.92 & \underline{98.52} & \underline{8.69} & 94.02 & 34.70 & 98.46 & 7.73 & 69.24 & 90.10 \\
 &  & $||\b z||_1$ & 47.74 & 95.45 & 70.81 & 66.69 & 53.48 & 87.82 & 73.95 & 78.05 & 73.89 & 66.14 & 58.42 & 89.18 & 86.16 & 51.18 & 99.61 & 1.36 & 50.14 & 93.10 \\
 &  & Residual & 50.18 & 94.86 & 85.59 & 50.02 & 80.17 & 68.38 & 76.76 & 80.65 & \underline{97.67} & 10.99 & 67.60 & 98.55 & 95.43 & 25.34 & 99.95 & 0.00 & \underline{81.57} & \underline{66.20} \\
 &  & Max Logit & 77.25 & 71.07 & 90.24 & 41.72 & 88.11 & 59.64 & 91.87 & 48.90 & 87.08 & 55.70 & \underline{99.04} & \underline{4.64} & 96.25 & 25.98 & 98.79 & 6.47 & 70.57 & 90.70 \\
 &  & Energy & 74.68 & 77.15 & 89.41 & 45.23 & 85.86 & 68.88 & 89.27 & 59.78 & 85.85 & 61.61 & \textbf{99.19} & \textbf{3.17} & 96.56 & 23.82 & 98.83 & 6.83 & 70.33 & 92.55 \\
 &  & Gradnorm & 64.64 & 88.00 & 84.85 & 46.15 & 73.53 & 76.05 & 87.99 & 53.65 & 85.04 & 50.85 & 94.56 & 29.39 & 95.94 & 22.56 & 99.82 & 0.67 & 57.05 & 89.85 \\
 &  & ViM & 70.30 & 86.87 & \textbf{94.95} & 25.61 & 92.08 & 41.79 & 91.68 & 47.40 & \textbf{99.17} & \textbf{3.39} & 95.59 & 29.88 & \textbf{99.30} & \textbf{1.26} & \textbf{100.00} & 0.00 & \textbf{86.80} & \textbf{55.55} \\
 &  & Mahal & 56.82 & 93.95 & 89.62 & 46.82 & 86.43 & 61.39 & 85.09 & 73.14 & \underline{98.19} & \underline{9.19} & 77.36 & 98.76 & 96.09 & 22.40 & 99.88 & 0.00 & \underline{84.28} & \underline{62.85} \\
\bottomrule
\end{tabular}
}
    \resizebox{\textwidth}{!}{
    \begin{tabular}{lllbbaallllllllllllll}
\toprule
 &  &  & \multicolumn{2}{c}{\textbf{ID\xmark}} & \multicolumn{2}{c}{\textbf{OOD mean}} & \multicolumn{2}{c}{\textbf{Openimage-O}} & \multicolumn{2}{c}{\textbf{iNaturalist}} & \multicolumn{2}{c}{\textbf{Textures}} & \multicolumn{2}{c}{\textbf{Colonoscopy}} & \multicolumn{2}{c}{\textbf{Colorectal}} & \multicolumn{2}{c}{\textbf{Noise}} & \multicolumn{2}{c}{\textbf{ImageNet-O}} \\
\textbf{Model} &  & \textbf{Method} & AUROC$\uparrow$ & FPR@95$\downarrow$ & AUROC$\uparrow$ & FPR@95$\downarrow$ & AUROC$\uparrow$ & FPR@95$\downarrow$ & AUROC$\uparrow$ & FPR@95$\downarrow$ & AUROC$\uparrow$ & FPR@95$\downarrow$ & AUROC$\uparrow$ & FPR@95$\downarrow$ & AUROC$\uparrow$ & FPR@95$\downarrow$ & AUROC$\uparrow$ & FPR@95$\downarrow$ & AUROC$\uparrow$ & FPR@95$\downarrow$ \\

\cmidrule(r){1-5} \cmidrule{6-21}
\multirow[c]{16}{*}{\begin{sideways}\shortstack[l]{\textbf{DenseNet-121} \\ ID \%Error: 25.58}\end{sideways}} & \multirow[c]{6}{*}{\begin{sideways}\textbf{SIRC}\end{sideways}} & (MSP,$||\b z||_1$) & \underline{85.99} & \underline{63.14} & 89.52 & 33.01 & 90.93 & 39.50 & 95.36 & 21.61 & 89.65 & 37.34 & 96.79 & 17.15 & 96.06 & 20.94 & 99.74 & 1.10 & 58.10 & 93.45 \\
 &  & (MSP,Res.) & \underline{85.97} & \underline{63.33} & 90.05 & 31.62 & 91.17 & 38.58 & 94.08 & 27.50 & 93.38 & 22.93 & 96.18 & 19.71 & 95.51 & 23.60 & 99.67 & 0.60 & 60.35 & 88.45 \\
 &  & (DR,$||\b z||_1$) & 85.77 & 64.51 & 90.00 & 30.09 & 91.55 & 36.00 & \underline{95.98} & \underline{17.81} & 90.32 & 33.12 & 97.10 & 14.22 & 96.88 & 15.68 & 99.79 & 0.83 & 58.36 & 93.00 \\
 &  & (DR,Res.) & 85.72 & 65.09 & 90.43 & \underline{28.80} & 91.72 & \underline{35.28} & 94.50 & 24.33 & \underline{93.85} & \underline{19.42} & 96.30 & 16.79 & 96.21 & 17.90 & 99.62 & 0.54 & 60.83 & 87.35 \\
 &  & ($-\mathcal{H}$,$||\b z||_1$) & 84.90 & 67.31 & 90.83 & \underline{28.44} & \underline{92.41} & \underline{34.47} & \textbf{96.52} & \textbf{16.28} & 91.05 & 32.85 & \underline{97.89} & \underline{9.86} & \underline{97.79} & 12.36 & 99.83 & 0.68 & 60.31 & 92.60 \\
 &  & ($-\mathcal{H}$,Res.) & 84.85 & 67.87 & \underline{91.46} & \textbf{26.46} & \textbf{92.64} & \textbf{34.09} & 95.67 & 20.33 & \underline{94.42} & \underline{19.07} & 97.45 & 11.37 & 97.56 & \underline{12.30} & 99.79 & \underline{0.39} & 62.71 & 87.70 \\
  \cmidrule(r){2-5} \cmidrule{6-21}
 & \multirow[c]{10}{*}{} & MSP & \textbf{86.11} & \textbf{62.67} & 88.81 & 36.77 & 90.26 & 43.08 & 94.26 & 27.56 & 88.31 & 43.72 & 96.90 & 17.10 & 94.44 & 30.72 & 99.55 & 1.69 & 57.97 & 93.55 \\
 &  & DOCTOR & 85.93 & 63.43 & 89.28 & 34.17 & 90.82 & 39.93 & 94.83 & 23.95 & 88.85 & 41.01 & 97.33 & 13.52 & 95.24 & 25.94 & 99.64 & 1.37 & 58.23 & 93.45 \\
 &  & $-\mathcal{H}$ & 84.97 & 66.76 & 90.39 & 30.68 & 91.91 & 37.18 & 95.83 & \underline{19.56} & 90.08 & 37.56 & \underline{97.95} & \underline{9.42} & 97.00 & 17.20 & 99.76 & 1.15 & 60.17 & 92.70 \\
 &  & $||\b z||_1$ & 47.53 & 94.93 & 78.50 & 53.82 & 69.94 & 70.15 & 89.06 & 39.14 & 84.61 & 49.73 & 58.85 & 89.84 & 92.89 & 34.90 & 99.88 & 0.49 & 54.29 & 92.50 \\
 &  & Residual & 51.52 & 94.26 & 71.96 & 66.47 & 69.78 & 78.27 & 61.14 & 93.61 & 90.21 & 33.64 & 37.94 & 99.37 & 75.49 & 70.74 & 97.32 & 14.40 & \underline{71.83} & \underline{75.25} \\
 &  & Max Logit & 77.97 & 71.35 & \textbf{91.62} & 28.87 & \underline{92.19} & 38.48 & \underline{96.07} & 20.57 & 91.59 & 34.32 & \textbf{98.20} & \textbf{8.77} & \underline{98.62} & \underline{6.48} & \underline{99.89} & 0.49 & 64.77 & 92.95 \\
 &  & Energy & 76.13 & 75.77 & \underline{91.47} & 30.02 & 91.54 & 42.66 & 95.60 & 23.50 & 91.39 & 35.43 & 97.87 & 11.18 & \textbf{98.86} & \textbf{5.02} & \underline{99.91} & 0.39 & 65.12 & 91.95 \\
 &  & Gradnorm & 55.44 & 92.10 & 85.31 & 42.04 & 78.97 & 58.55 & 93.87 & 25.24 & 89.62 & 37.81 & 81.08 & 68.36 & 97.63 & 13.10 & \textbf{99.96} & \underline{0.09} & 56.04 & 91.15 \\
 &  & ViM & 70.16 & 88.53 & 89.58 & 47.81 & 88.40 & 56.49 & 88.74 & 66.34 & \textbf{96.64} & \textbf{17.69} & 82.83 & 89.17 & 95.19 & 31.76 & 99.57 & \textbf{0.01} & \underline{75.66} & \underline{73.20} \\
 &  & Mahal & 57.28 & 94.10 & 68.90 & 81.55 & 69.02 & 86.67 & 49.94 & 97.79 & 82.79 & 55.43 & 66.51 & 96.97 & 58.34 & 96.34 & 75.13 & 68.35 & \textbf{80.53} & \textbf{69.30} \\
\bottomrule
\end{tabular}
}
\end{table}

\subsection{Varying $\alpha$ and $\beta$}
\begin{figure}
     \centering
    \includegraphics[width=\linewidth]{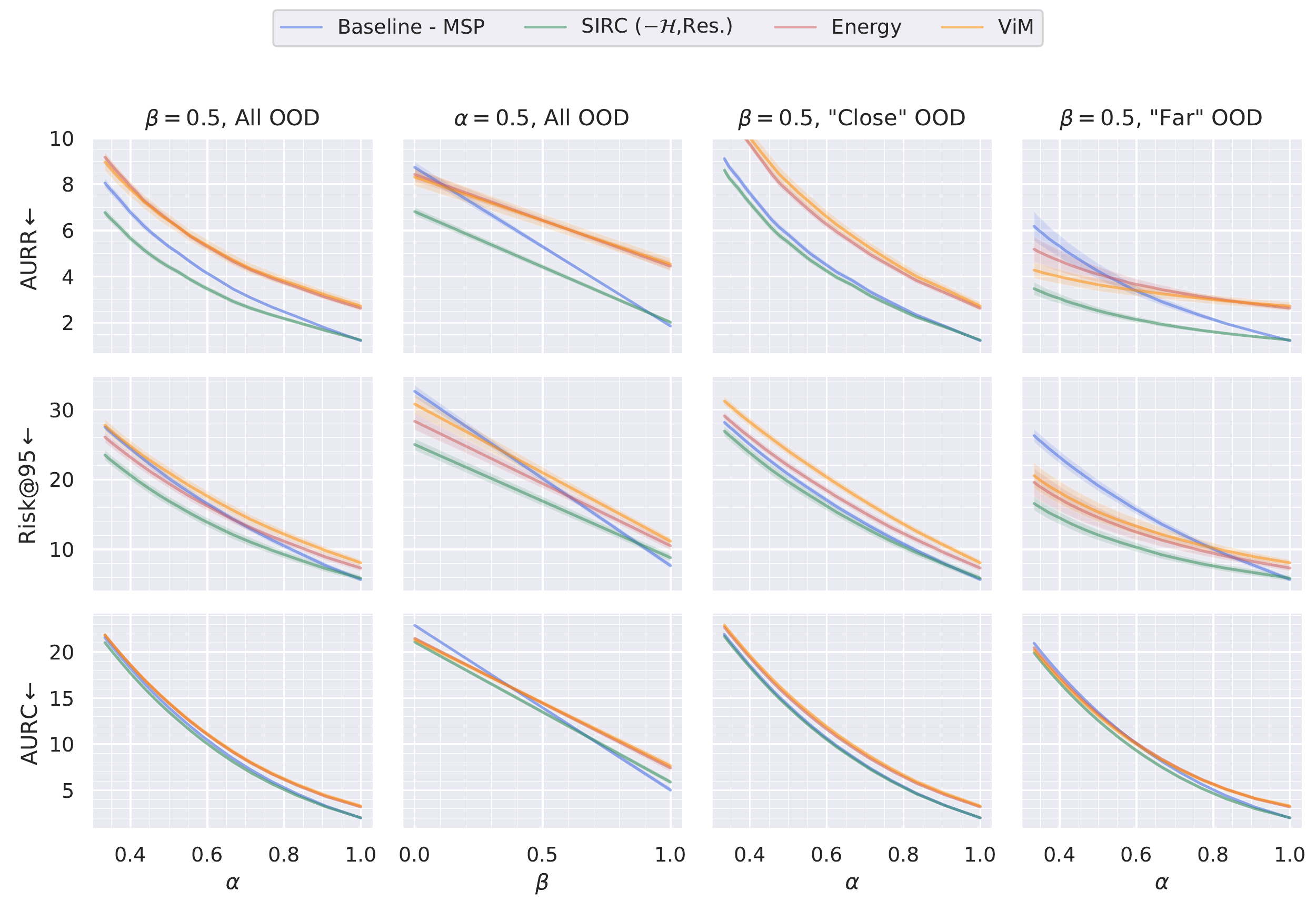}
    \caption{Varying $\alpha$ and $\beta$ for ResNet-50 (ImageNet-200) (values $\times 10^2$).}
    \label{fig:vary1}
\end{figure}

\begin{figure}
    \centering
    \includegraphics[width=\linewidth]{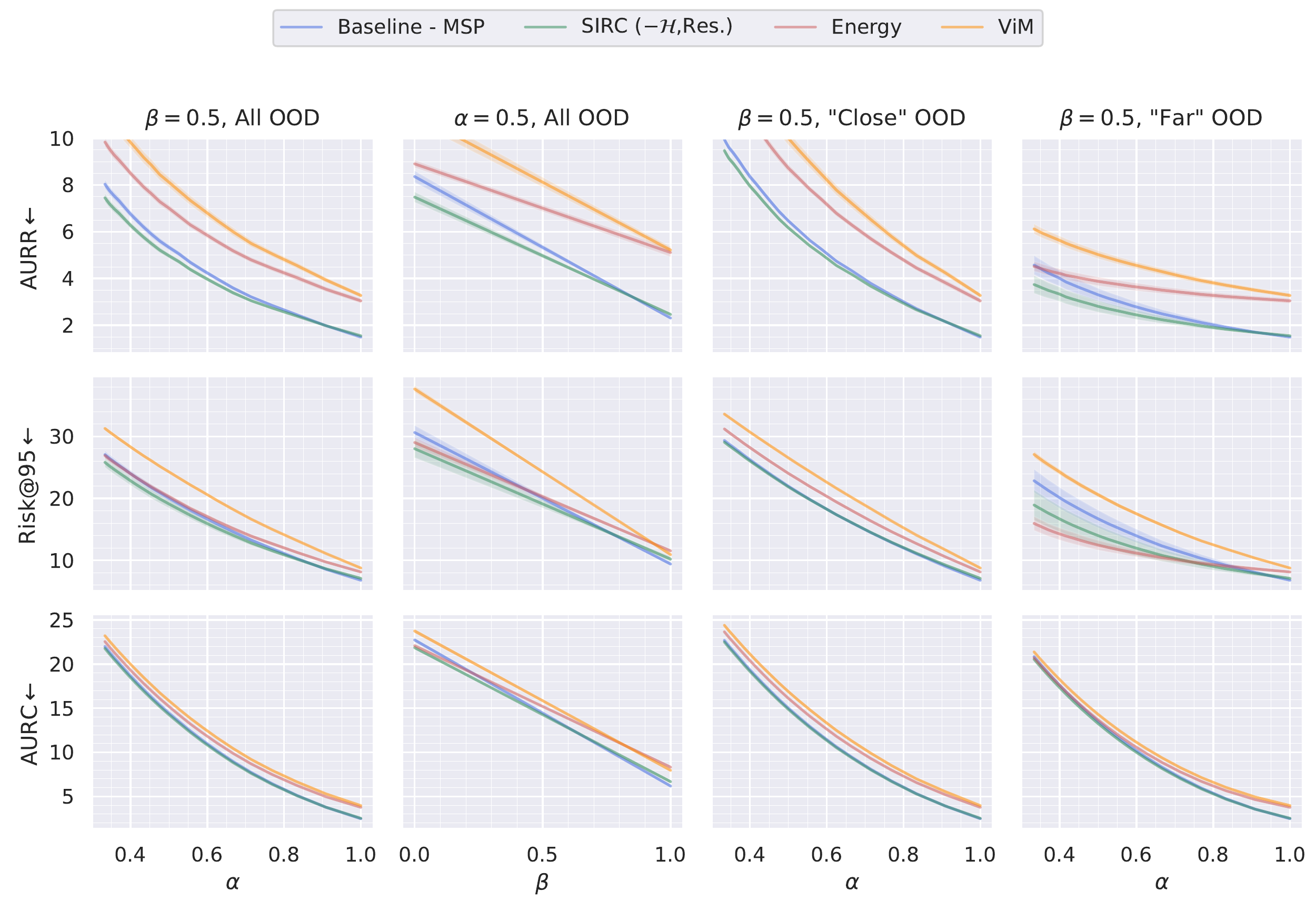}
    \caption{Varying $\alpha$ and $\beta$ for MobileNetV2 (ImageNet-200) (values $\times 10^2$).}
    \label{fig:vary2}
   \end{figure}

\begin{figure}
    \centering
    \includegraphics[width=\linewidth]{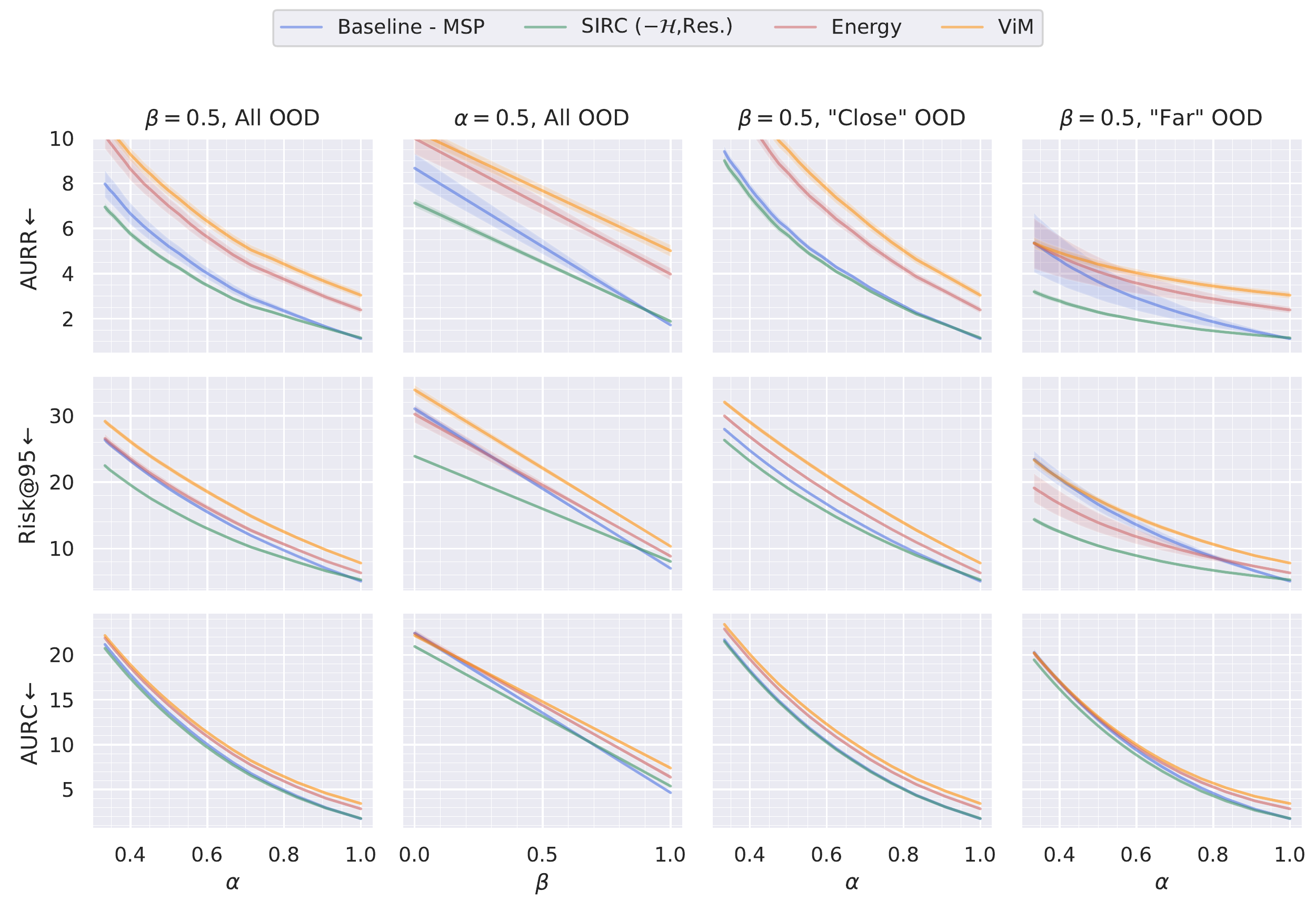}
    \caption{Varying $\alpha$ and $\beta$ for DenseNet-121 (ImageNet-200) (values $\times 10^2$).}
    \label{fig:vary3}
   \end{figure}

We plot versions of Fig. \ref{fig:vary} for all 3 ImageNet-200 architectures (Figs. \ref{fig:vary1} to \ref{fig:vary3}). We also present the mean$\pm$std. The ability of SIRC to perform consistently better than the baseline generalises across the 3 different CNN architectures. We note that differences in AURC are harder to distinguish, due to the metric considering the proportion of all input data accepted, rather than just the recall of  ID\cmark. The behaviour, however, is similar to AURR in terms of relative performance to the baseline, so we omit AURC from the main results.

\subsection{SCOD vs OOD Detection}
\begin{figure}
     \centering
    \includegraphics[width=\linewidth]{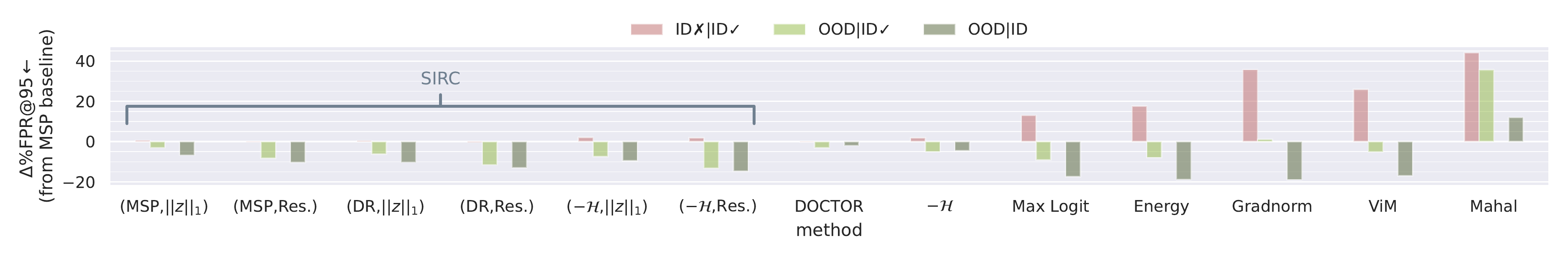}
    \caption{ResNet-50 (ImageNet-200), comparing the change in \%FPR@95 relative to the MSP baseline for different detection methods and data groups.}
    \label{fig:bar1}
\end{figure}

\begin{figure}
     \centering
    \includegraphics[width=\linewidth]{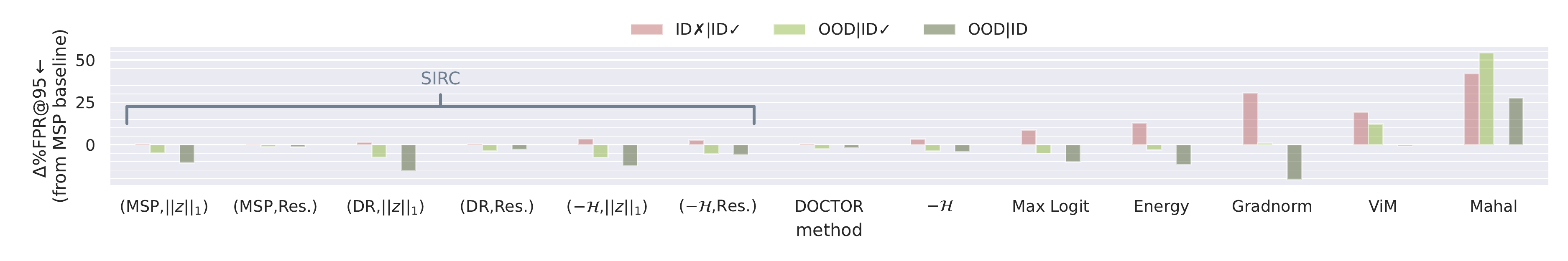}
    \caption{MobileNetV2 (ImageNet-200), comparing the change in \%FPR@95 relative to the MSP baseline for different detection methods and data groups.}
    \label{fig:bar2}
\end{figure}

\begin{figure}
     \centering
    \includegraphics[width=\linewidth]{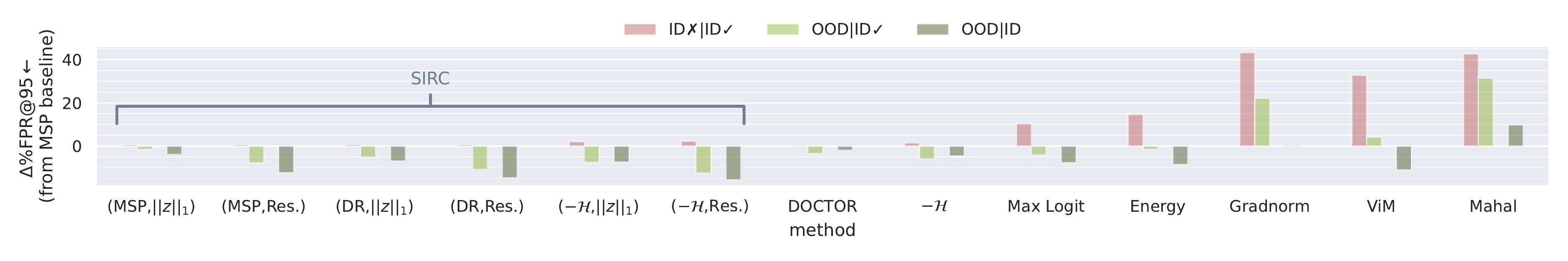}
    \caption{DenseNet-121 (ImageNet-200), comparing the change in \%FPR@95 relative to the MSP baseline for different detection methods and data groups.}
    \label{fig:bar3}
\end{figure}

\begin{figure}
     \centering
    \includegraphics[width=\linewidth]{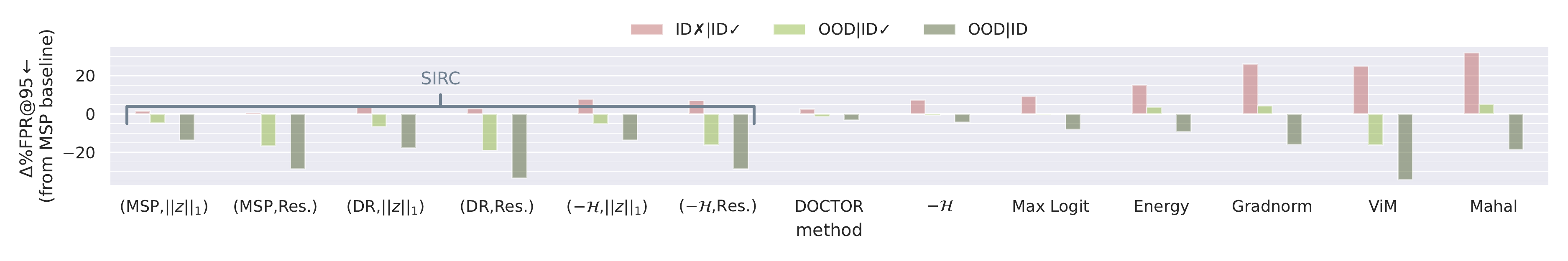}
    \caption{ResNetV2-101 (ImageNet-1k), comparing the change in \%FPR@95 relative to the MSP baseline for different detection methods and data groups.}
    \label{fig:bar4}
\end{figure}

\begin{figure}
     \centering
    \includegraphics[width=\linewidth]{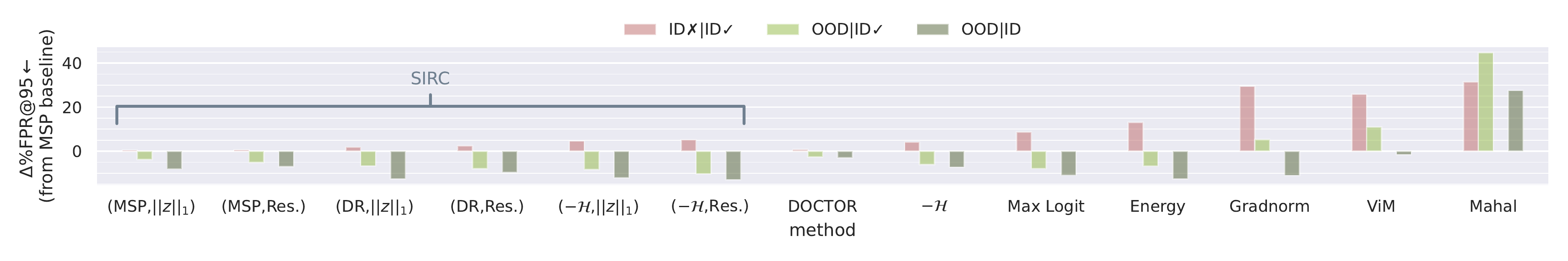}
    \caption{DenseNet-121 (ImageNet-1k), comparing the change in \%FPR@95 relative to the MSP baseline for different detection methods and data groups.}
    \label{fig:barlast}
\end{figure}

Similar to the previous section we include versions of Fig.\ref{fig:ood_scod} for all architectures and confidence scores (Figs. \ref{fig:bar1} to \ref{fig:barlast}).  The behaviour is as discussed in Section \ref{sc_comp}, with methods designed for OOD detection achieving gains over the baseline for OOD detection by sacrificing their ability to separate ID\xmark|ID\cmark.
\subsection{Plotting $S_2$ against $S_1$}
\begin{figure}
     \centering
    \includegraphics[width=\linewidth]{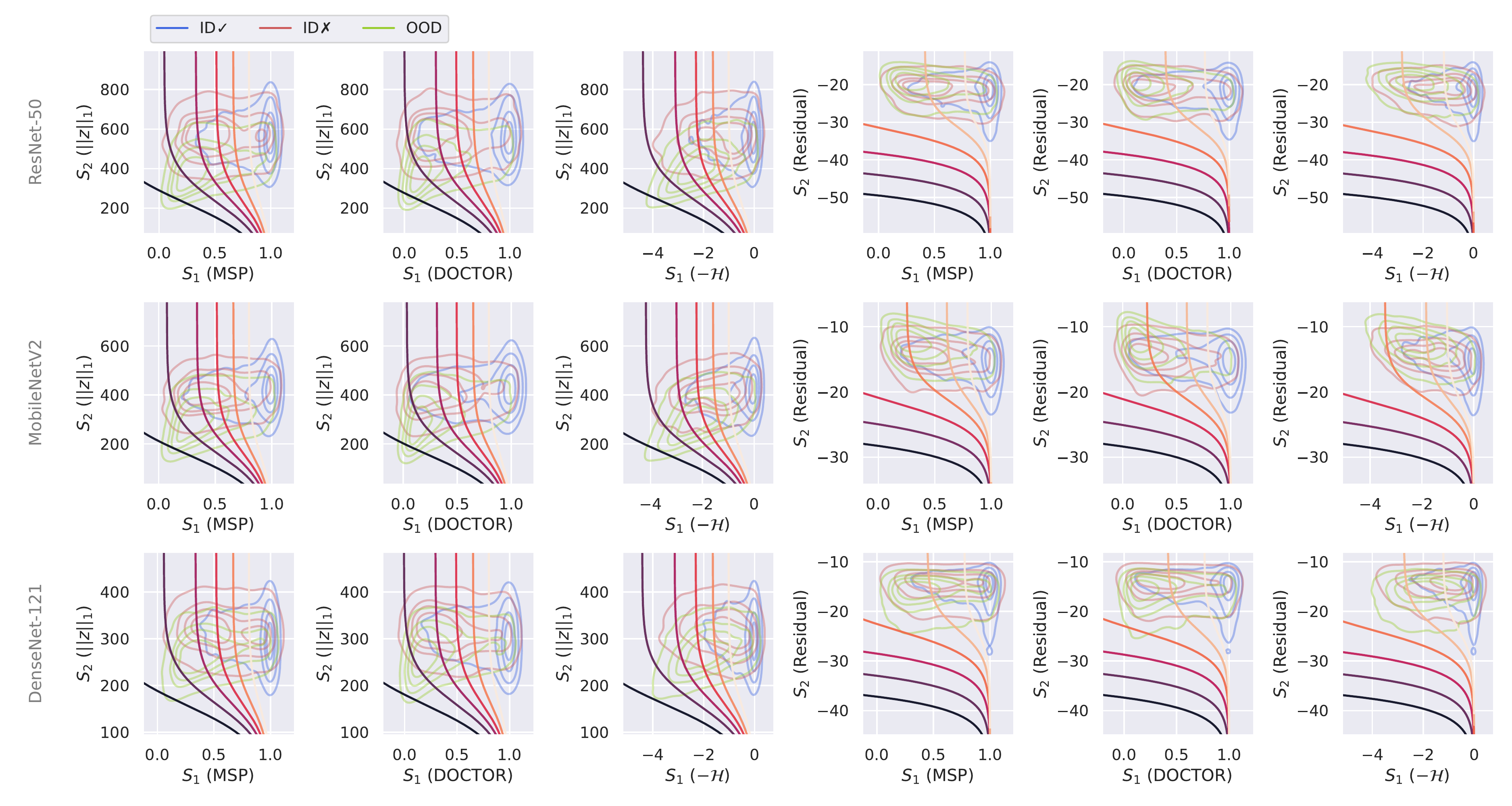}
    \caption{SIRC combinations on the $S_1,S_2$-plane, ID: ImageNet-200, OOD: iNaturalist.}
    \label{fig:comb1}
\end{figure}

\begin{figure}
    \centering
    \includegraphics[width=\linewidth]{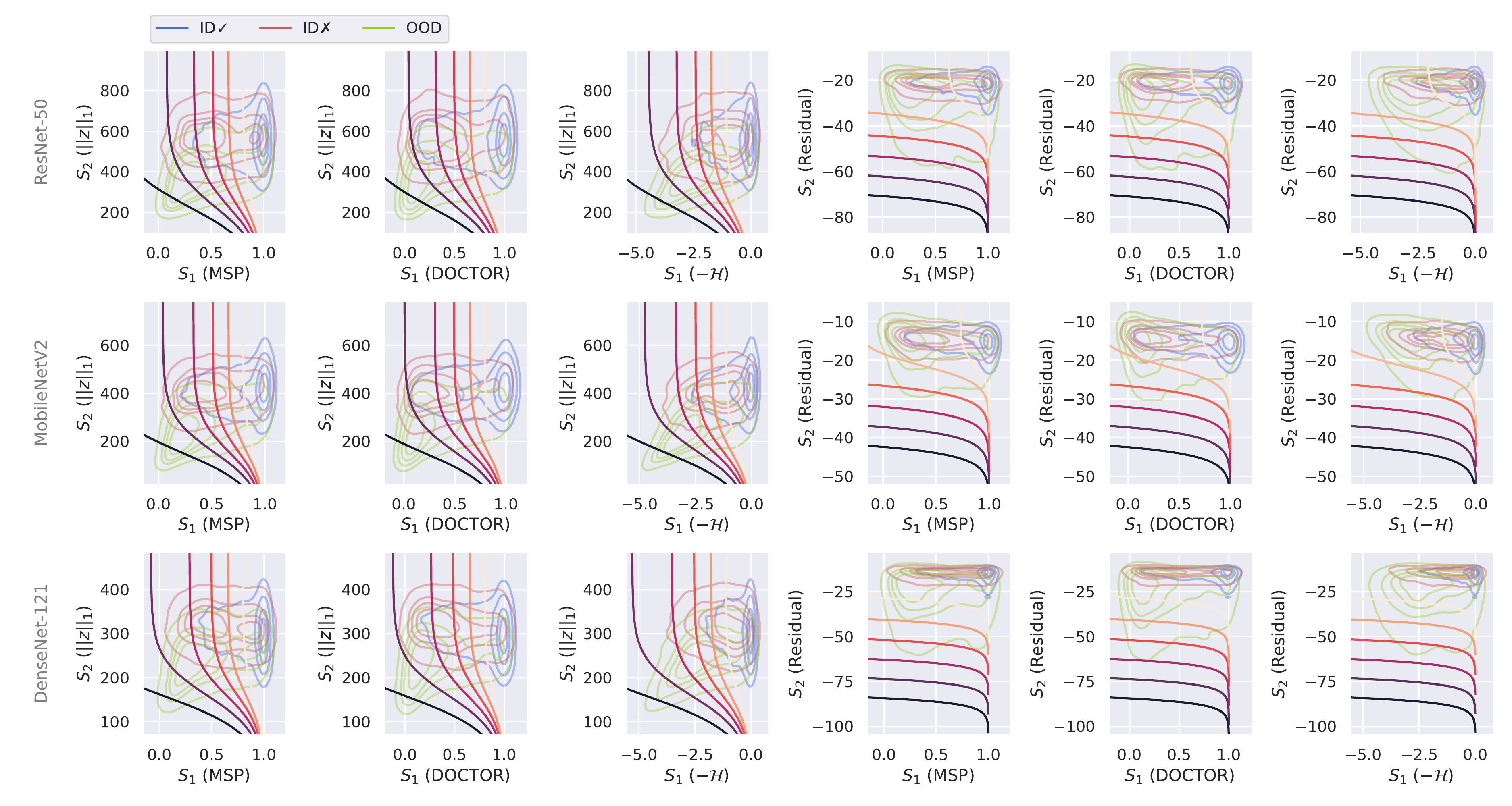}
    \caption{SIRC combinations on the $S_1,S_2$-plane, ID: ImageNet-200, OOD: Textures.}
    \label{fig:comb2}
   \end{figure}

\begin{figure}
     \centering
    \includegraphics[width=\linewidth]{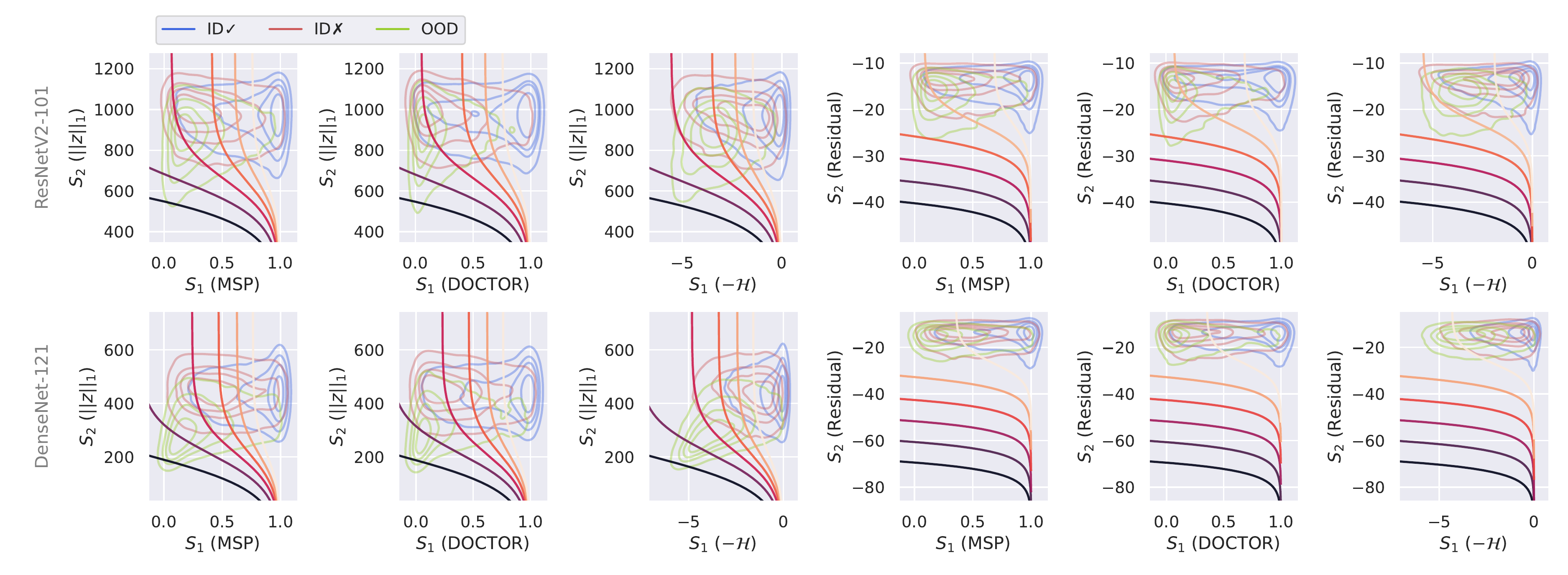}
    \caption{SIRC combinations on the $S_1,S_2$-plane, ID ImageNet-1k, OOD: iNaturalist.}
    \label{fig:comb3}
\end{figure}

\begin{figure}
    \centering
    \includegraphics[width=\linewidth]{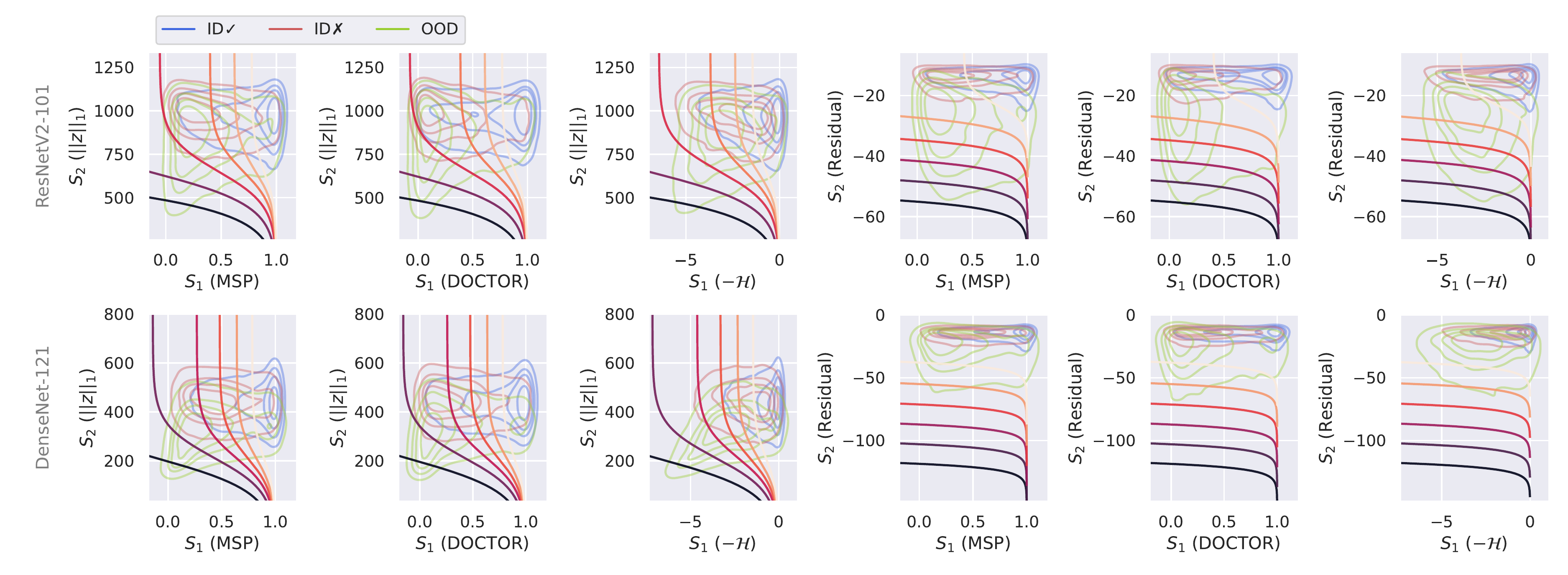}
    \caption{SIRC combinations on the $S_1,S_2$-plane, ID ImageNet-1k, OOD: Textures.}
    \label{fig:comblast}
   \end{figure}

In a similar vein to Figure \ref{fig:combs}, we plot different SIRC combinations on the $S_1,S_2$-plane for different experimental configurations (Figs. \ref{fig:comb1} to \ref{fig:comblast}). If there are multiple training runs, we plot the distributions corresponding to the outputs of the 1st run. Decision contours corresponding to the default parameter setting for SIRC are also overlayed. We note that the inconsistency of Residual can be observed here, where in some cases the OOD distribution is much lower than ID, whilst in others, there is almost complete overlap. In the case of MobileNetV2 on iNaturalist it is in fact higher for OOD than ID, although the nature of SIRC means that it is robust to such $S_2$ failure (as discussed in Section \ref{indep}).



\end{document}